\documentclass[runningheads]{llncs}

\usepackage{eccv}

\usepackage{eccvabbrv}

\usepackage{graphicx}
\usepackage{booktabs}
\usepackage{pifont}
\usepackage{makecell}
\usepackage[table]{xcolor}
\usepackage[colorlinks=true, urlcolor=blue]{hyperref} % 1. 加载宏包并设置链接样式
\usepackage[accsupp]{axessibility}  % Improves PDF readability for those with disabilities.

\usepackage{hyperref}

\begin{document}

\title{Tac2Real: Reliable and GPU Visuotactile Simulation for Online Reinforcement Learning and Zero-Shot Real-World Deployment} 

\titlerunning{Tac2Real}

\author{Ningyu Yan\inst{1,2}$^*$ \and
Shuai Wang\inst{1}$^*$ \and
Xing Shen\inst{1} \and
Hui Wang\inst{1,4} \and
Hanqing Wang\inst{1}$^\dagger$ \and
Yang Xiang \inst{2,3}$^\dagger$ \and
Jiangmiao Pang\inst{1}$^\dagger$
}

\authorrunning{N. Yan et al.}

\institute{Shanghai Artificial Intelligence Laboratory. \and
The Hong Kong University of Science and Technology. \and Algorithms of Machine Learning and Autonomous Driving Research Lab, HKUST
Shenzhen-Hong Kong Collaborative Innovation Research Institute. \and Shanghai Jiao Tong University
}

\maketitle
\begin{flushleft}
\let\thefootnote\relax\footnotetext{$^*$Equal contribution. $^\dagger$ Corresponding authors.}
\end{flushleft}

\begin{abstract}
% Visuotactile sensors play significant roles in contact-rich manipulation tasks for robots. However, policy learning in simulation with tactile sensing remains challenging—particularly in online reinforcement learning (RL), where balancing the fidelity and efficiency of visuotactile simulation is critical. We present Tac2Real, a visuotactile simulation library supporting online RL training, which integrates Preconditioned Nonlinear Conjugate Gradient Incremental Potential Contact (PNCG-IPC) method and multi clusters \& GPUs high-throughput parallel simulation mechanism. Tac2Real can generate very realistic visuotactile images and the displacement field of markers with very fast speed, and Tac2Real can be easily integrated into any physics engine, such as IsaacLab, to perform RL training. Furthermore, Tac2Real also summarizes a systematic solution for narrowing the sim2real gap from systematic to random aspects in order to guarantee the successful sim2real transfer. Finally, we evaluate the sim2real performance of Tac2Real in 2 contact-rich tasks, peg insertion and nut threading, both with a random rotation of the held objects. The results shows a high success rate, demonstrating the effectiveness of Tac2Real.
Visuotactile sensors are indispensable for contact-rich robotic manipulation tasks. 
However, policy learning with tactile feedback in simulation, especially for online reinforcement learning (RL), remains a critical challenge, as it demands a delicate balance between physics fidelity and computational efficiency.
To address this challenge, we present Tac2Real, a lightweight visuotactile simulation framework designed to enable efficient online RL training.
Tac2Real integrates the Preconditioned Nonlinear Conjugate Gradient Incremental Potential Contact (PN\allowbreak CG-IPC) method with a multi-node, multi-GPU high-throughput parallel simulation architecture, which can generate marker displacement fields at interactive rates.
Meanwhile, we propose a systematic approach, TacAlign, to narrow both structured and stochastic sources of domain gap, ensuring a reliable zero-shot sim-to-real transfer.
We further evaluate Tac2Real on the contact-rich peg insertion task.
The zero-shot transfer results achieve a high success rate in the real-world scenario, verifying the effectiveness and robustness of our framework. The project page is: \href{https://ningyurichard.github.io/tac2real-project-page/}{https://ningyurichard.github.io/tac2real-project-page/}

\end{abstract}

\section{Introduction}
\label{sec:intro}

\begin{figure}[htbp]
  \centering
  \includegraphics[width=1.0\textwidth]{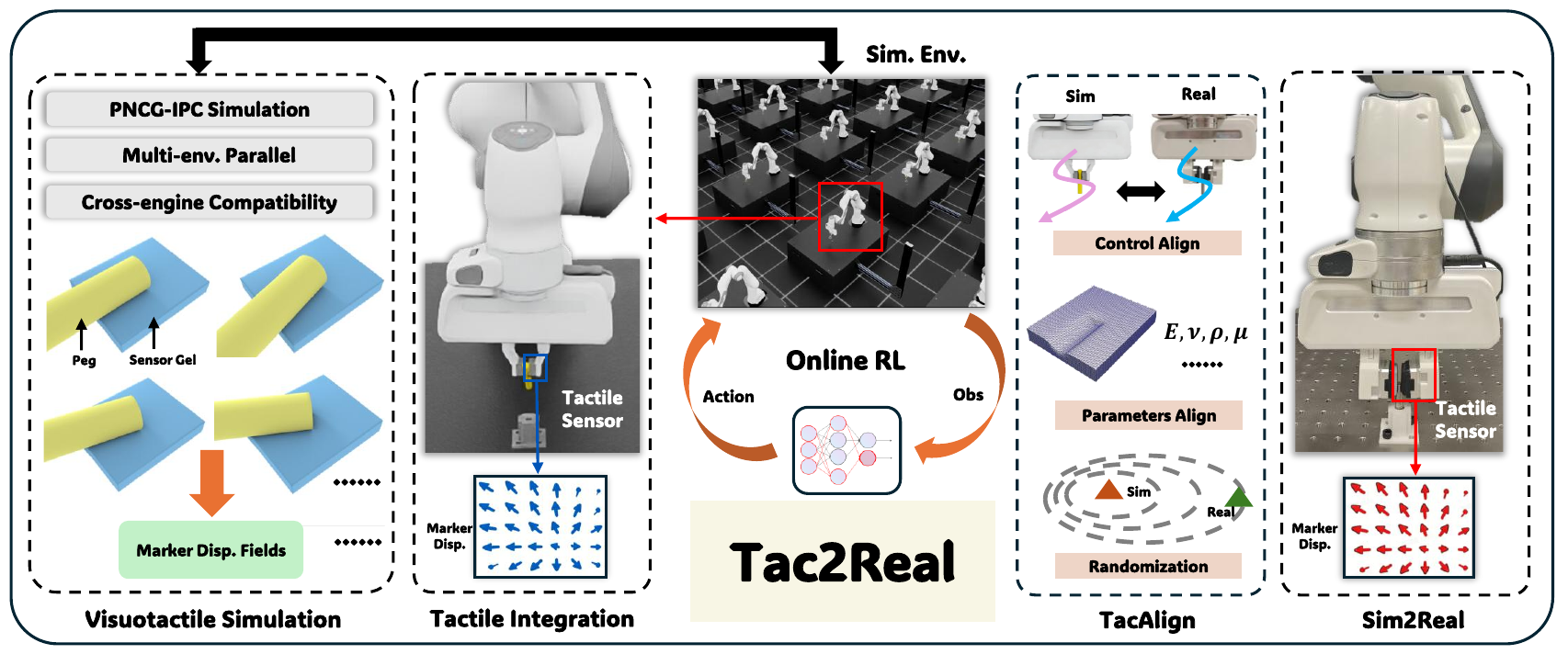}  % 替换为你的 PDF 文件名
  \caption{Tac2Real Framework.Tac2Real contains a visuotactile simulator, which can be seamlessly integrated to existing physics engines in a highly parallel way. Tac2real supports online large-scale RL training. Through the systematic approach introduced in TacAlign framework, the sim-to-real gap will be significantly reduced, and Tac2real guarantees a reliable real-world policy deployment. }
  \label{fig:framework}
\end{figure}

Tactile sensing is a cornerstone of human dexterity and has long motivated the design of robotic manipulation systems \cite{tegin2005tactile, yousef2011tactile, wang2024multimodal}. 
Our sense of touch enables us to perceive contact forces, surface textures, and local geometries, which is essential for contact-rich behaviors such as compliant grasping, precise assembly, and hand manipulation, especially under degraded or ambiguous visual feedback \cite{yu2023mimictouch, xue2025reactive, zhao2025polytouch}. 
In robotics, vision-based tactile sensors (VBTSs) such as the GelSight family have become a widely adopted sensing modality \cite{abad2020visuotactile, yuan2015measurement, yuan2016estimating}. 
By capturing high-resolution tactile images and marker displacement fields, VBTSs translate low-level contact physics into rich perceptual signals that align naturally with modern computer vision and learning-based frameworks. 

However, integrating VBTS into data-driven policy training represents a key challenge for online RL.
The fundamental limitation lies in efficiently synthesizing high-fidelity simulated tactile signals. 
While some penalty-based \cite{wang2022tacto, akinola2025tacsl} approaches have been successfully applied to scalable tactile data generation, they fall short of modeling soft deformation and multi-phase contact dynamics that underpin realistic VBTS sensing.
In contrast, high‑fidelity physics-based alternatives such as material point methods (MPM) \cite{de2020material,jiang2016material,bardenhagen2004generalized,sulsky2004implicit} can capture deformations with greater accuracy, yet often struggle with shear response and numerical stability under large deformations \cite{zhang2025nitsche, ma2014new, huang2011contact}, making them unsuitable for complex contact-rich tasks. 
Furthermore, few existing simulation pipelines are architected for multi-GPU acceleration, restricting their use in large-scale parallelization for online RL. 
Against this backdrop, the zero-shot sim-to-real transfer \cite{zhao2026high, qureshi2025splatsim, gu2024humanoid, zhong2024tactgen} for tactile-based RL remains largely unresolved, with limited successful demonstrations on real hardware.

In this work, we present \textbf{Tac2Real}, a high-fidelity visuotactile simulation framework designed to address these limitations, as shown in \cref{fig:framework}. 
At its core, Tac2Real uses the Preconditioned Nonlinear Conjugate Gradient Incremental Potential Contact (PNCG-IPC) solver \cite{shen2024preconditioned}, which builds typical IPC method \cite{li2020incremental, huang2024gipc, lan2021medial, ferguson2023high} and guarantees physically consistent elastomer simulation with robust contact handling. 
By deploying this solver on a multi-node, multi-GPU architecture, we achieve the simulation throughput that enables large-scale online RL. 
To mitigate uncertainties and high-dimension cost introduced by optical rendering, Tac2Real directly outputs the marker displacement field as tactile representation and is designed as a lightweight plugin, allowing seamless integration with mainstream robot simulators such as Isaac Lab \cite{mittal2025isaaclab} and MuJoCo \cite{todorov2012mujoco}.

Beyond simulation, we identify that reliable sim-to-real transfer requires a structured strategy to reduce the reality gap \cite{hietala2021closing, josifovski2022analysis, blanco2024benchmarking, zhao2026closing}. 
This gap includes both structured discrepancies, such as mismatched robot dynamics, material parameters, and contact models, and stochastic variations from unmodeled noise and environmental uncertainty. 
To close this gap, we introduce \textbf{TacAlign}, a unified, general calibration pipeline for tactile tasks. It consists of four sequential stages:
(i) low-level robot controller calibration via trajectory alignment;
(ii) identification of material parameters through both baseline and task-engaged tests;
(iii) task‑specific fine‑tuning of friction and contact parameters; and
(iv) domain randomization during RL training.

We evaluate Tac2Real on the challenging blind peg-in-hole task with random initial orientations, a benchmark that exemplifies the inherent difficulties of contact-rich manipulation. 
Experiments show that policies trained entirely in simulation transfer zero-shot to a real robot with high success rates, outperforming baselines built on TacSL \cite{akinola2025tacsl} and Tacchi \cite{chen2023tacchi}. 
These results highlight the importance of both high-fidelity tactile simulation and reality-gap mitigation for the real-world deployment of tactile-based RL policies.

To summarize, this work makes the following contributions:
\begin{itemize}
\item[\textbullet] \textbf{Tac2Real}, a high‑performance visuotactile simulation framework based on PNCG‑IPC, which supports multi-GPU parallelization and enables seamless integration with mainstream physics engines for RL.
\item[\textbullet] \textbf{TacAlign}, a systematic methodology for both reducing structured and stochastic sim‑to‑real gaps in tactile contact‑rich manipulations.
\item[\textbullet] Experimental validations on real-world peg insertion tasks, demonstrating superior zero-shot transfer performance and robustness over existing state-of-the-art approaches.
\end{itemize}
We believe that Tac2Real provides a robust foundation for scaling up tactile‑based policy learning, and takes a significant step toward to the long‑standing goal of reliable, real‑world dexterous manipulations.

\section{Related Work}
Research on visuotactile simulation has evolved through three distinct stages: standalone sensor-object modeling, closed-loop policy generation (with or without learning) and sim-to-real trial, and holistic system integration and real-world deployment. Initially, methods like Tacchi~\cite{chen2023tacchi} and Tacchi 2.0~\cite{sun2025tacchi} utilized the Material Point Method (MPM) to accurately model elastomer deformation and generate tactile images, though they often struggled with friction modeling and numerical stability; this prompted a shift toward the more robust Incremental Potential Contact (IPC) method, leading to simulators like TacIPC~\cite{du2024tacipc}. The second stage extended these capabilities to policy generation and sim-to-real transfer through differentiable optimization (Difftactile~\cite{si2024difftactile}) or reinforcement learning integration~\cite{sun2025soft, chen2024general}, yet these approaches typically lacked a unified, physically consistent simulation of the entire robot–object–sensor system. Currently, the third class achieves holistic integration within scalable environments like Isaac Gym \cite{makoviychuk2021isaac}, where frameworks such as TacSL~\cite{akinola2025tacsl}, TacFlex~\cite{zhang2025tacflex}, and Taccel~\cite{li2025taccel} combine penalty-based \cite{xu2021end, xu2023efficient}, Finite Element Method (FEM) \cite{dhatt2012finite, zienkiewicz1977finite}, or IPC-based tactile models with robot dynamics to support large-scale parallel training and reliable real-world deployment (summarized in \cref{tab:related works}).

Our proposed Tac2Real utilizes the PNCG-IPC method~\cite{shen2024preconditioned}, an enhanced IPC solver specifically designed to shorten convergence time.
By leveraging multi-node and multi-GPU acceleration and seamless integration with existing physics engines, Tac2Real enables large-scale online RL while maintaining high visuotactile fidelity. 
Also, Tac2Real incorporates a systematic strategy for mitigating sim-to-real gaps. We validate the robustness of our framework through zero-shot real-world deployment.
\begin{table}[tb]
  \centering
  % --- 关键修改：在这里添加字体大小命令 ---
  \scriptsize   % 选项: \small (稍大), \footnotesize (推荐), \scriptsize (最小)
  \renewcommand{\theadfont}{\bfseries\scriptsize}
  \caption{Comparison of proposed Tac2Real with existing methods. S, O, and R denote sensor, object, and robot, respectively. The ranking in
contact fidelity is based on suitability for online tactile RL, considering
deformation realism, frictional contact handling, and numerical robustness. BC is behavior cloning.}
  \label{tab:related works}
  % 可选：调整行高，防止文字换行后太拥挤 (1.2 倍行高)
  \renewcommand{\arraystretch}{1.1} 
  \begin{tabular}{@{}l c c c c c@{}}
    \toprule
    % 确保你已经在 preamble 添加了 \usepackage{makecell}
    \textbf{Method} & \textbf{Scene} & \textbf{\thead{Sim. Method/\\Contact Fidelity}} & \textbf{\thead{Tactile \\ Representation}} & \textbf{\thead{Policy\\Learning}} & \textbf{Sim2Real} \\
    \midrule
    Tacchi \cite{chen2023tacchi} & S-O & MPM / medium & RGB & $\times$ & $\times$ \\
    Tacchi 2.0 \cite{sun2025tacchi} & S-O & MPM / medium & Markers & $\times$ & $\times$ \\
    TacIPC \cite{du2024tacipc} & S-O & IPC / high & RGB & $\times$ & $\times$ \\
    Difftactile \cite{si2024difftactile} & S-O & MPM+FEM / medium & Markers/RGB & $\times$ & $\checkmark$ \\
    Sun \etal~\cite{sun2025soft} & S-O & MPM / medium & Markers/RGB & RL & $\checkmark$ \\
    Chen \etal~\cite{chen2024general} & S-O & IPC / high & Markers & RL & $\checkmark$ \\
    TacSL \cite{akinola2025tacsl} & S-O-R & Non-physics / low & Markers/RGB &RL/BC & $\checkmark$ \\
    TacFlex \cite{zhang2025tacflex} & S-O-R & FEM / high & Markers/RGB &BC & $\checkmark$ \\
    Taccel \cite{li2025taccel} & S-O-R & IPC / high & Markers/RGB &$\times$ & $\checkmark$ \\
    \textbf{Tac2Real (ours)} & \textbf{S-O-R} & \textbf{IPC / high} & \textbf{Markers} &\textbf{RL} & \textbf{$\checkmark$} \\
    \bottomrule
  \end{tabular}
\end{table}

\section{Tac2Real Simulation Framework}
\subsection{PNCG-IPC Method}

Our tactile simulation is built upon the PNCG-IPC solver~\cite{shen2024preconditioned}, which strikes an effective balance between physical accuracy and computational efficiency for contact simulation.

IPC formulates elastodynamic contact simulation as a variational optimization problem. Using implicit Euler integration, the positions $\mathbf{x}^{t+1}$ at each time step are obtained by minimizing:
% \begin{equation}
%   E(\mathbf{x}) = \tfrac{1}{2}(\mathbf{x}-\hat{\mathbf{x}})^\top \mathbf{M}(\mathbf{x}-\hat{\mathbf{x}}) + h^2\Psi(\mathbf{x}) + \kappa \textstyle\sum_{k\in C} b(d_k(\mathbf{x})) + D(\mathbf{x}),
% \end{equation}
\begin{equation}
  E(\mathbf{x}) = \tfrac{1}{2}(\mathbf{x}-\hat{\mathbf{x}})^\top \mathbf{M}(\mathbf{x}-\hat{\mathbf{x}}) + h^2\Psi(\mathbf{x}) + B(\mathbf{x}) + D(\mathbf{x}),
\end{equation}
% where the four terms represent the inertia potential, the hyperelastic energy $\Psi$, and the log-barrier contact potential $B(\mathbf{x})$ that approximates the inequality constraints $h_l(\mathbf{x})\geq 0$, respectively. The standard IPC pipeline employs Newton's method with CCD-based line search, which yields high per-iteration accuracy but is computationally expensive and difficult to parallelize on GPUs.
where the four terms represent the inertia potential, the hyperelastic energy $\Psi$, and the log-barrier contact potential $B(\mathbf{x})$ and frictional potential $D(\mathbf{x})$, respectively. More details about PNCG-IPC are provided in supplementary material. The standard IPC pipeline employs Newton's method with continuous collision detection (CCD) based line search, which yields high per-iteration accuracy but is computationally expensive and difficult to parallelize on GPUs.
PNCG-IPC replaces Newton's method with the nonlinear conjugate gradient algorithm, which directly solves the nonlinear optimization without assembling or factorizing the Hessian matrix. The algorithm only requires gradient evaluations, diagonal Hessian entries, and vector-vector dot products—all highly GPU-parallelizable operations. A CCD-free line search strategy further eliminates the expensive collision detection per iteration by deriving an analytical step-size upper bound.

This design philosophy trades per-iteration accuracy for dramatically faster iteration throughput: while each conjugate gradient step is less precise than a Newton step, the extremely low per-iteration cost on GPUs allows PNCG-IPC to converge to sufficient accuracy within only tens of iterations. For tactile simulation where visually plausible and physically consistent deformation matters more than machine-precision convergence, this trade-off is highly favorable.

Moreover, PNCG-IPC is implemented as a lightweight solver based on the Taichi programming language~\cite{hu2019taichi}, which compiles to high-performance GPU kernels from concise Python code. This compact implementation makes it straightforward to integrate PNCG-IPC as a plug-in module into diverse simulation pipelines, as detailed in \cref{sect:integration}.

\subsection{Tactile Representation}
\begin{figure}[htbp]
  \centering
  \includegraphics[width=1.0\textwidth]{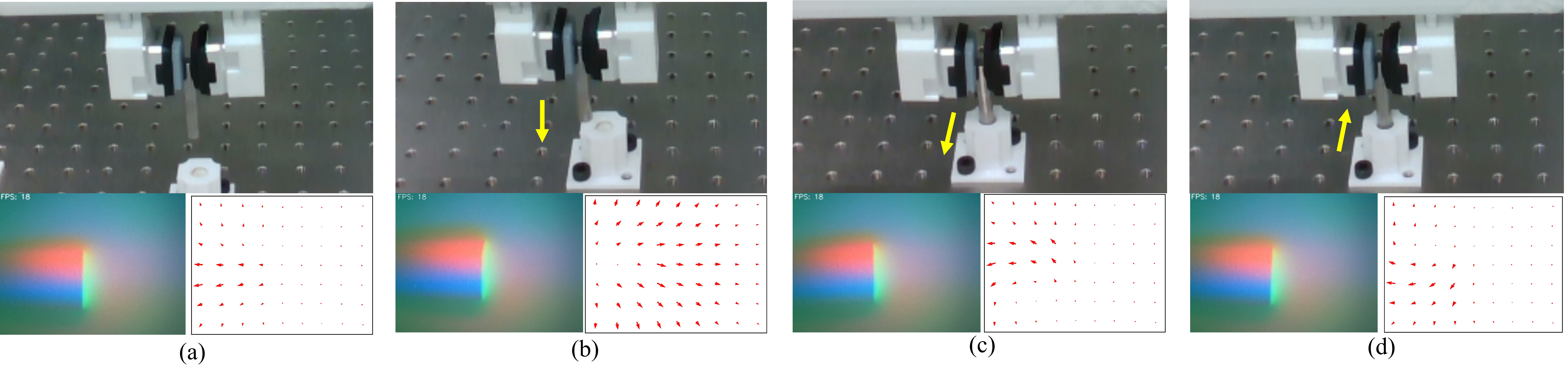}  % 替换为你的 PDF 文件名
  \caption{Tactile feedback under different contact modes. (a) stationary state; (b) press-down state; (c) move forward state; (d) move backward state.}
  \label{fig:contact}
\end{figure}

In this work, we mainly focus on simulating the tactile representations of Gelsight Mini sensor, which can output either an RGB tactile image (320$\times$240) or a marker displacement field (9$\times$7) depending on the sensor gel type.
Marker displacement fields excel at capturing diverse contact states, which is critical for robots contact-rich manipulation tasks, whereas RGB tactile images are useful for texture identification. 
To select the appropriate representation, we conduct a simple collision test. 
We mounted tactile sensors on the gripper of a Franka Panda robot, then grasped a peg to perform various contact interactions with sockets. 
We recorded both RGB outputs and 2D marker displacement fields under static, press-down, move-forward, and move-backward conditions.
As shown in \cref{fig:contact}, the marker displacement fields exhibit significant variations across different collision states, while RGB images show only subtle differences.
This superior sensitivity of markers to distinct contact modes, combined with their lower-dimensional representation, may prove crucial for large-scale RL training by enhancing regularity and efficiency.
In our simulation, a mapping between marker and initial IPC mesh nodes can be constructed using the k-nearest neighbor method; then, the displacements of marker are interpolated weightedly from mesh node positions in the deformed state.

\subsection{Integrating to Physics Engine} \label{sect:integration}
The integration of our tactile simulation into existing physics engine for RL training follows a plugin-based architecture.
As shown in \cref{fig:sim frame}, the RL agent first sends the action command, triggering one simulation step in the physics engine.
We then extract the relative position and orientation between the grasped object and the tactile sensor during this step, and compute the corresponding linear and angular velocities which are used in the following tactile simulation. 
Finally, output markers displacement fields, together with the baseline observation in the physics engine forms the combined observation which is fed back to the RL agent to continue training. This integrating method has following main properties:
\vspace{-0.5cm}
\begin{figure}[htbp]
  \centering
  \includegraphics[width=1.0\textwidth]{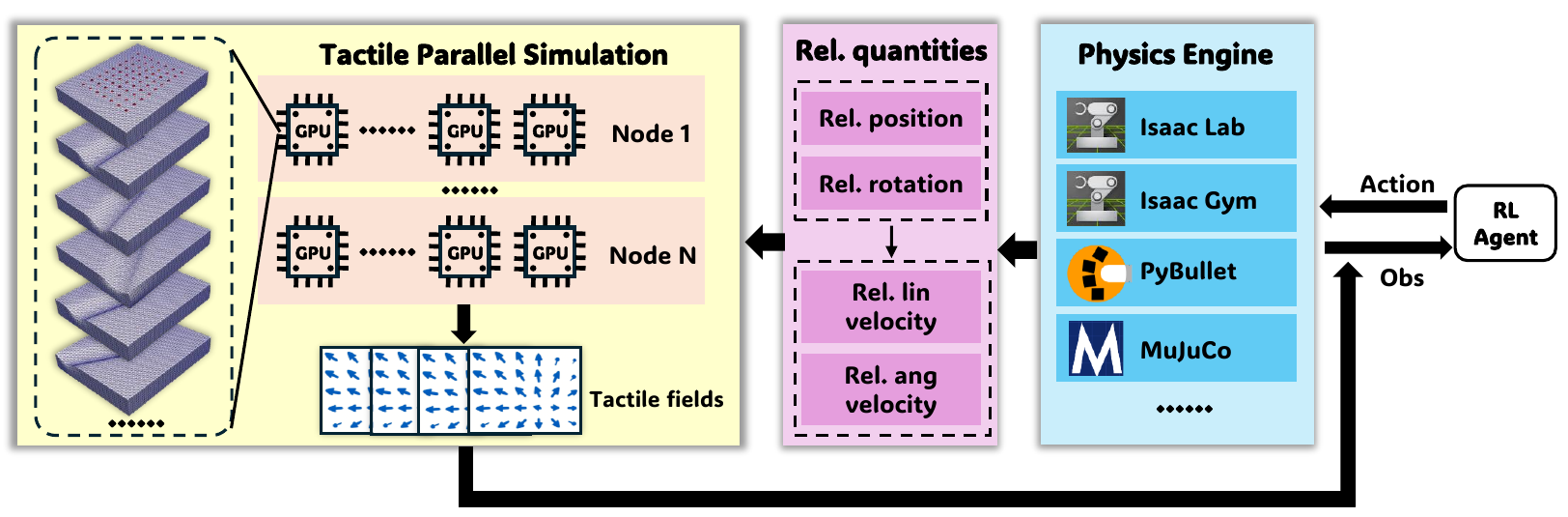}  % 替换为你的 PDF 文件名
  \caption{Tactile Simulation Framework. Our tactile simulation is an external interface outside the physics engine. Receiving relative quantities from physics environments, Tac2Real starts tactile simulation in a multi-node multi-GPU paralle way with each GPU assigned multi environments' tactile tasks. }
  \label{fig:sim frame}
\end{figure}
\vspace{-0.5cm}

\textbf{Cross-engine compatibility}. 
Our tactile simulation only requires relative physical quantities between the sensor elastomer and the interacting objects, so this interface can be embedded within the environment file layered on top of the physic engine, for example, the \_get\_observations() function in Isaac Lab \cite{mittal2025isaaclab}, or play\_steps\_rnn() function in rl-games library \cite{rl-games2021}. 

\textbf{Multi-node Multi-GPU parallelization}. 
Benefiting from the parallelizability of the PNCG-IPC and Taichi language, \textbf{Tac2Real} leverages multi-node, multi-GPU parallelization to enable robust and efficient RL training.
Specifically, we construct a Ray \cite{moritz2018ray} cluster containing multi nodes, each equipped with multiple GPUs. 
A Ray-wrapped tactile simulation class is instantiated for each GPU, which manages tactile simulation for a set of environments.
During the online RL roll-out process, the simulation function within the Ray-wrapped class is invoked iteratively: it takes the relative pose quantities as input and returns the marker displacement fields after simulation. 
Finally, Ray-based distributed communication across multiple nodes is used to gather simulation results from all GPUs.

\section{TacAlign to Narrow Sim2real Gap} \label{sec:tacalign}
\begin{figure}[htbp]
  \centering
  \includegraphics[width=1\textwidth]{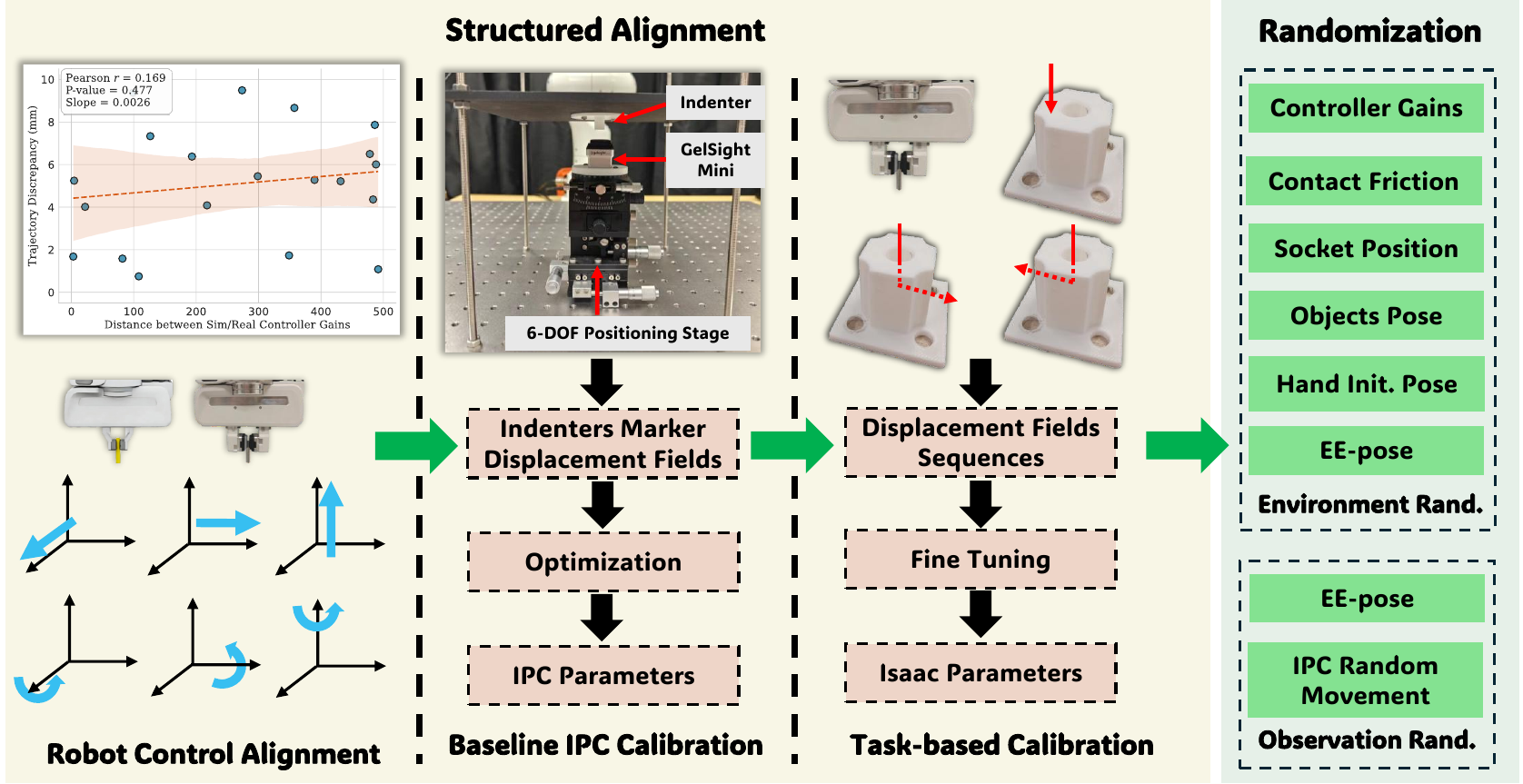}  % 替换为你的 PDF 文件名
  \caption{TacAlign Framework. TacAlign narrows the sim-to-real gap from both structured and stochastic perspectives.}
  \label{fig:tacalign}
\end{figure}

To systematically mitigate the sim-to-real gap, we propose \textbf{TacAlign} within the \textbf{Tac2Real} framework, addressing both structured and stochastic discrepancies, as illustrated in \cref{fig:tacalign}.
\vspace{-0.5cm}
\vspace{-1cm}
% \textbf{Robot controll alignment}. In our work, both Isaac Lab and real Franka use impedance-controller. The target applied force $F^{targ}$ on the end-effector is:
% \begin{equation}
%   F^{targ} = k_p (p^{targ}(a) - p^{ee}) - k_d v^{ee},
%   \label{eq:important}
% \end{equation}
% where $k_p, k_d$ are controller gains and $k_d = 2 \sqrt{k_p}$; $F^{targ}$ is calculated by the linear combination of the mismatch of target ee-pose $p^{targ}$, which is a function of applied action $a$, and current ee-pose $p^{ee}$, along with the velocity of end-effector. Furthurmore, we can formulate the ee-pose trajectory for sim and real as $x_{t+1}^{\text{sim}} = f_{\text{sim}}(x_t, a_t; k_p^{sim})$ and $x_{t+1}^{\text{real}} = f_{\text{real}}(x_t, a_t; k_p^{real})$, respectively. In real-world deployment, 
\subsection{Robot Control Alignment} \label{sec:tacalign_control}
In our framework, the Franka robot in simulation and real world both adopt Cartesian impedance control. 
The target force applied at the end-effector is given by:$ \mathbf{F}^{targ} = \mathbf{k_p} * (\mathbf{p}^{targ}(\mathbf{a}) - \mathbf{p}^{ee}) - \mathbf{k_d} * \mathbf{v}^{ee},$ where $\mathbf{k_p}$ and $\mathbf{k_d}$ denote the proportional and damping gains respectively, with $\mathbf{k_d} = 2\sqrt{\mathbf{k_p}}\in \mathbb{R}^6$ to ensure critical damping. 
Here, $\mathbf{p}^{targ}(\mathbf{a})\in \mathbb{R}^6$ is the target end-effector pose determined by the action $\mathbf{a}$, $\mathbf{p}^{ee}$ is the current end-effector pose, and $\mathbf{v}^{ee}$ is the end-effector velocity. * is the element-wise multiplication.

By this controller, the end-effector trajectories in simulation and reality can be expressed as:
\begin{equation}
\mathbf{x}_{t+1}^{\text{sim}} = f_{\text{sim}}(\mathbf{x}_t, \mathbf{a}_t; \mathbf{k_p}^{\text{sim}}), 
\quad
\mathbf{x}_{t+1}^{\text{real}} = f_{\text{real}}(\mathbf{x}_t, \mathbf{a}_t; \mathbf{k_p}^{\text{real}}),
\end{equation}
where $f_{\text{sim}}$ and $f_{\text{real}}$ represent the closed-loop system dynamics induced by the impedance controller.
A intuition is that reducing the discrepancy between $\mathbf{k_p}^{\text{sim}}$ and $\mathbf{k_p}^{\text{real}}$ would synchronously reduce the sim-to-real trajectory gap. 
In practice, however, this relationship is highly non-linear and non-monotonic due to unmodeled dynamics, actuator delays, frictional effects, and contact nonlinearities.
We empirically validate this phenomenon in the upper left part of \cref{fig:tacalign}, where we randomly choose 20 sim-real $\mathbf{k_p}$ pairs and visualize the trajectory discrepancy. 
Notably, the trajectory discrepancy does not necessarily decrease as $|\mathbf{k_p}^{\text{sim}} - \mathbf{k_p}^{\text{real}}|$ becomes smaller, and exhibits no significant linear correlation.
This observation suggests that simply matching controller parameters is insufficient to minimize the sim-to-real gap; instead, trajectory-level alignment is required.
We therefore formulate control alignment as the problem of minimizing trajectory discrepancy:$\mathcal{D}(\mathbf{k_p}^{\text{sim}}, \mathbf{k_p}^{\text{real}})=\frac{1}{T} \sum_{t=1}^{T}\left\| \mathbf{x}_t^{\text{sim}} - \mathbf{x}_t^{\text{real}} \right\|^2$, where the discrepancy is computed over six canonical end-effector motions (single-axis translation and rotation) as shown in \cref{fig:tacalign} .
Rather than treating either domain as ground truth, we jointly calibrate both simulation and reality using alternating minimization:
\begin{align}
\mathbf{k_p}^{\text{sim}, (k+1)} 
&= \arg\min_{\mathbf{k_p}^{\text{sim}}} 
\mathcal{D}(\mathbf{k_p}^{\text{sim}}, \mathbf{k_p}^{\text{real}, (k)}), \\
\mathbf{k_p}^{\text{real}, (k+1)} 
&= \arg\min_{\mathbf{k_p}^{\text{real}}} 
\mathcal{D}(\mathbf{k_p}^{\text{sim}, (k+1)}, \mathbf{k_p}^{\text{real}}).
\end{align}
This bidirectional procedure iteratively reduces trajectory divergence and alleviates controller-induced sim-to-real mismatches. 

\subsection{Baseline IPC Calibration}
Material parameters in PNCG-IPC simulation, specifically Young's modulus $E$, Poisson's ratio $\nu$, density $\rho$, and friction coefficient $\mu$, play a critical role in reducing the sim-to-real gap.
To ensure that the mechanical response of the elastomer in tactile simulation closely matches physical reality, we conduct a baseline calibration experiment.
First, we use the apparatus shown in \cref{fig:tacalign}, which contains a 6-DOF positioning stage, a GelSight Mini sensor and several 3D-printed indenters, to record sequences of marker displacement fields under different indenters and deformation modes. 
Let $\mathbf{u}^{\text{real}}_{k,i}$ and $\mathbf{u}^{\text{sim}}_{k,i}(\boldsymbol{\theta})$ denote the marker displacement fields for the $i$ indenter at frame $k$ in the real and simulated settings respectively, where $\boldsymbol{\theta} = [E, \nu, \rho, \mu]^\top$ is the vector of material parameters.
We then replicate these deformation modes in the IPC simulator and define the mean squared error (MSE) between simulated and real displacement fields as the loss function $\mathcal{L}(\boldsymbol{\theta})$:
\begin{equation}
    \mathcal{L}(\boldsymbol{\theta}) = \frac{1}{K \cdot N} \sum_{k=1}^{K} \sum_{i=1}^{N} \left\| \mathbf{u}^{\text{sim}}_{k,i}(\boldsymbol{\theta}) - \mathbf{u}^{\text{real}}_{k,i} \right\|_2^2,
    \label{eq:loss_function}
\end{equation}
where $K$ is the total number of frames, and $N$ is the number of indenters. 
The calibration is formulated as an optimization problem to find the optimal parameter set $\boldsymbol{\theta}^*$ that minimizes this discrepancy:
\begin{equation}
    \boldsymbol{\theta}^* = \arg\min_{\boldsymbol{\theta}} \mathcal{L}(\boldsymbol{\theta}).
    \label{eq:optimization_problem}
\end{equation}
To solve this gradient-free nonlinear optimization efficiently, we adopt the Covariance Matrix Adaptation Evolution Strategy (CMA-ES)~\cite{hansen2006cma}.

\subsection{Task-based Calibration} \label{sec: task-based}
Given the complex interactions encountered in real-world contact-rich tasks, the baseline calibration alone may not capture the full range of sensor deformation.
Therefore, we further perform task-aware fine-tuning on simulation parameters of Isaac Lab, such as friction coefficient $\mu_{isaac}$ and contact stiffness $s_{isaac}$, under both static and dynamic contact conditions. 
Four representative contact states are considered: stationary grasping, press-down, forward and backward collisions as illustrated in \cref{fig:tacalign}.
We record the marker displacement field sequences for both simulation and reality, with the real measurements used as the ground-truth reference.
Then, we fine-tune the $\mu_{isaac}$ and $s_{isaac}$ until the MSE of displacement fields between the closest-matching frames falls below a predefined threshold.

\subsection{Randomization}
We model the sim-to-real discrepancy as parametric and observational uncertainty.
Accordingly, we randomize physical parameters (controller gains, friction), geometric configurations (object and socket poses), and sensing channels (end-effector pose noise and IPC perturbations).
This stochastic regularization encourages the policy to be invariant to low-level dynamics mismatches and contact variations.
Domain randomization complements our structured alignment by narrowing the stochastic gap that is difficult to resolve through deterministic calibration alone. Details are provided in supplementary material.

% \section{RL setting}
% We formulate peg-in-hole as an MDP $(\mathcal{S}, \mathcal{A}, \mathcal{T}, \mathcal{R})$.
% The state space $\mathcal{S}$ contains the entire state of the robot and environment, such as robot configuration, object poses, and tactile displacements, while the policy only receives partial observations. 

% \textbf{Observation.}
% To highlight the role of tactile sensing, the policy observes only:
% (i) end-effector pose $p^{ee}$,
% (ii) tactile marker displacement field $\mathbf{u}$, and
% (iii) previous action $a_{t-1}$.
% No object pose or camera information is provided.

% \textbf{Action.}
% The action represents an incremental Cartesian update of the end-effector pose, 
% which is executed by the impedance controller described in section \ref{sec:tacalign_control}.
% This formulation ensures smooth and physically consistent motion.

% \textbf{Reward.}
% The reward consists of three components:
% (1) a keypoint alignment term encouraging peg–socket alignment,
% (2) sparse bonuses for engagement and successful insertion, and
% (3) a contact-force penalty to discourage unstable interactions.

\section{Experiental Results}
\subsection{PNCG-IPC Simulation Results}
\begin{figure}[htbp]
  \centering
  \includegraphics[width=1.02\textwidth]{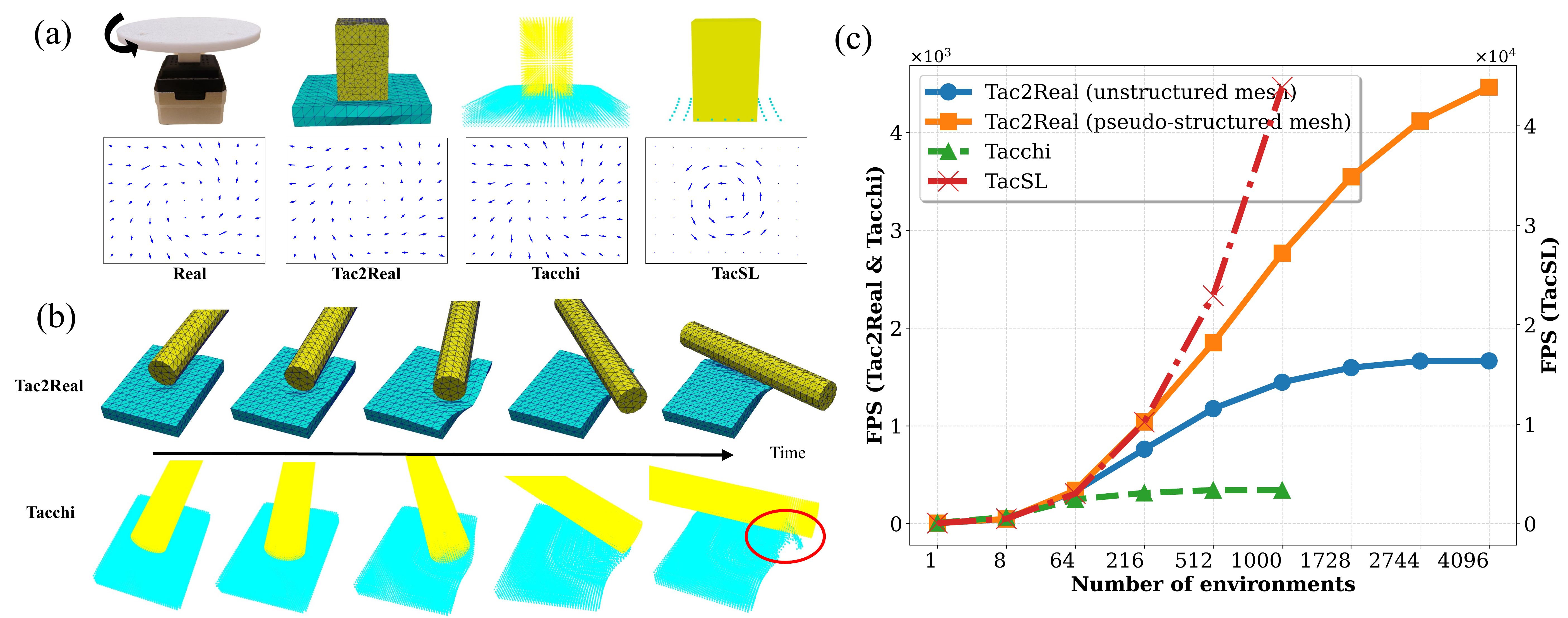}  % 替换为你的 PDF 文件名
  \caption{Comparison of different simulation methods. (a) Comparison among Tac2Real, Tacchi, TacSL and real reference in cube indentation test; (b) Comparison of Tac2Real and Tacchi in large rotation deformation; (c) Parallel performance comparison.}
  \label{fig:sim_compare}
\end{figure}

\noindent\textbf{Reliability.}
We evaluate PNCG-IPC (Tac2Real) against Tacchi and TacSL using a cube indenter rotation task. As shown in \cref{fig:sim_compare}(a), Tac2Real and Tacchi both yield realistic deformations consistent with real-world data, whereas TacSL exhibits significant discrepancies due to its non-physical, penalty-based approach that only models interpenetration regions. Under large rotational deformations (\cref{fig:sim_compare}(b)), Tacchi suffers from particle splashing and numerical instability caused by adhesion deficiencies. In contrast, Tac2Real robustly handles large rotations and slips, with the sensor reliably recovering its original state post-contact, demonstrating superior stability for contact-rich RL tasks. More simulation results and comparisons are provided in supplementary material.

\noindent\textbf{Parallel Performance.}
Benchmarked on a node with 16 RTX 4090 GPUs, Tac2Real achieves exceptional scalability (\cref{fig:sim_compare}(c)). At 4,096 environments, it reaches 4,465 FPS (pseudo-structured mesh) and 1,665 FPS (unstructured mesh), outperforming Tacchi. While TacSL achieves higher raw FPS by relying on simple SDF queries, it sacrifices physical fidelity. Tacchi's lower parallel efficiency stems from the overhead of constructing large background grids for multiple environments on a single GPU. Consequently, Tac2Real offers the optimal balance of high-throughput parallelism and physical accuracy required for online reinforcement learning.

\subsection{TacAlign Results}
\textbf{Control Alignment.} 
We employ the alternating optimization approach described above to search for an optimal combination of controller gains for both simulation and reality.
We start with the original value of $\mathbf{k_p}^{\text{sim}}=(100, 30)$ in Isaac Lab, which is an abbreviation for $(100, 100, 100, 30, 30, 30)$, and keep $\mathbf{k_p}^{\text{real}}$ to be the same.
Thresholds for the averaged trajectory discrepancy of translation $\bar{\mathcal{D}}_{\text{trans}}$ and rotation $\bar{\mathcal{D}}_{\text{rot}}$ along the 3 coordinate axis are set to 3~mm and $0.5^{\circ}$ respectively. 
As shown in \cref{tab:controller_alignment}, the initial averaged trajectory discrepancy is relatively large, with a translational mismatch of 11.11 mm recorded.
This large discrepancy makes it impossible to complete the peg insertion task, as the diameter of the socket hole in this task is only approximately 8~mm.
By the optimization, the trajectory discrepancy is reduced to 2.521~mm (translation) and $0.454^{\circ}$ (rotation).  
The corresponding optimized controller gains are $\mathbf{k_p}^{\text{sim}}=(600, 50)$ and $\mathbf{k_p}^{\text{real}}=(400, 20)$, which are fixed for all subsequent alignment procedures and RL training.
\setlength{\tabcolsep}{8pt}      % Increases horizontal space between columns (default is 6pt)
\begin{table}[h]
\centering
\scriptsize   % 选项: \small (稍大), \footnotesize (推荐), \scriptsize (最小)
\caption{Controller Alignment Performance.}
\label{tab:controller_alignment}
\begin{tabular}{cc|cc}
\toprule
\textbf{$\mathbf{k_p}^{\text{sim}}$} & \textbf{$\mathbf{k_p}^{\text{real}}$} & \textbf{$\bar{\mathcal{D}}_{\text{trans}}$} (mm) & \textbf{$\bar{\mathcal{D}}_{\text{rot}}$} (degree) \\
\midrule
(100, 30) & (100, 30) & 11.11 & 2.635 \\ % [4pt] adds extra space after this specific row
(300, 30) & (500, 40)& 7.381 & 1.091 \\
\textbf{(600, 50)} & \textbf{(400, 20)} & \textbf{2.521} & \textbf{0.454} \\
\bottomrule
\end{tabular}
\end{table}
% Reset spacing if needed for subsequent tables (optional if used locally)
% \renewcommand{\arraystretch}{1.0}
\setlength{\tabcolsep}{6pt}

\begin{figure}[htbp]
  \centering
  \includegraphics[width=1\textwidth]{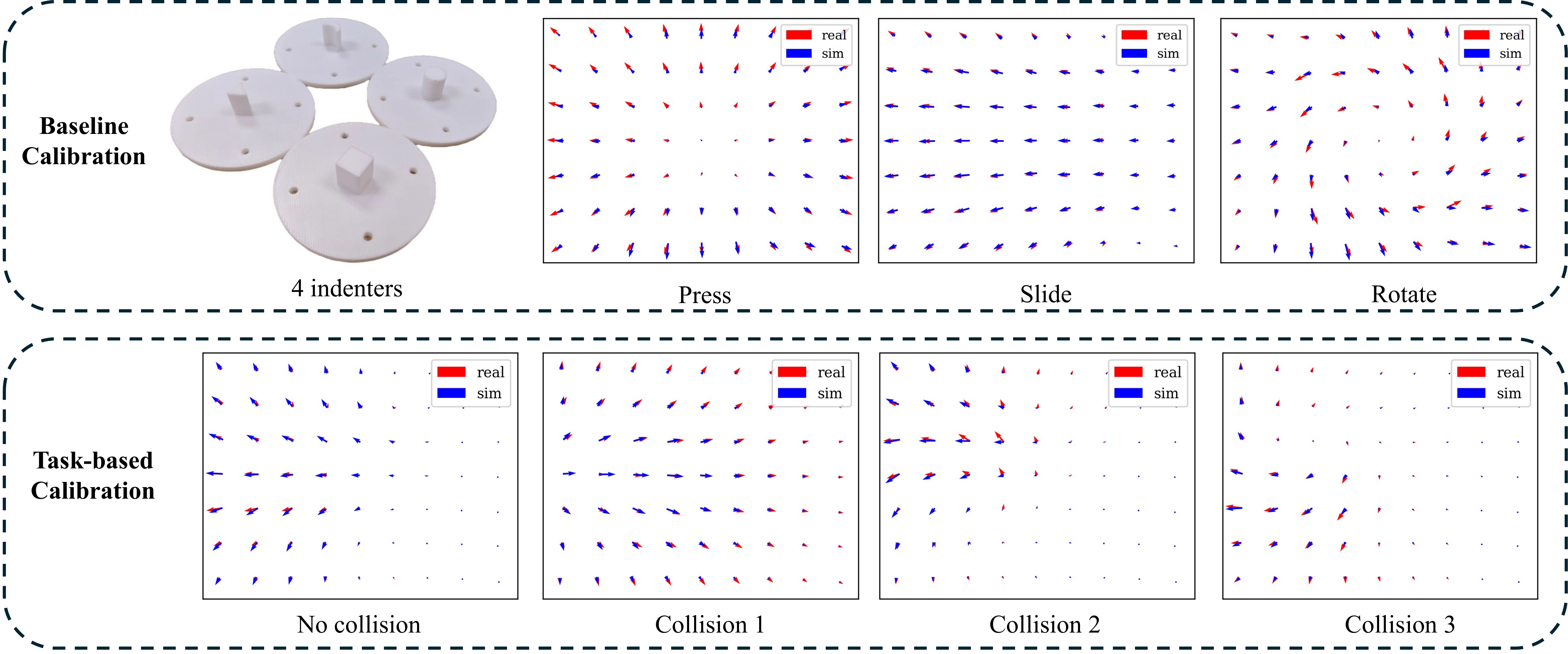}  % 替换为你的 PDF 文件名
  \caption{Simulation parameter calibration.}
  \label{fig:ipc cali}
\end{figure}

\noindent
\textbf{Baseline IPC Calibration Results.} 
Four indenters of distinct shapes, including the cube, cylinder, moon and triangle, are used in the baseline calibration. 
For each indenter, we performed three deformation interactions with the GelSight Mini sensor’s elastomer: pressing, sliding, and rotating.
The magnitudes of these three deformation modes are set to 1~mm, 1~mm and $2^{\circ}$, with step increments of 0.1~mm, 0.1~mm and $0.5^{\circ}$, respectively. 
We also constructed an identical calibration environment (comprising the sensor elastomer and indenters) using PNCG-IPC, simulating the entire deformation process to support the CMA-ES optimization.
The final result are presented in \cref{fig:ipc cali}, where only the results for the cube indenter at the final frame are illustrated.
As evident from the results, the mechanical behavior of the elastomer in simulation closely approximates that in the real world. More optimization information is provided in supplementary material.
\vspace{-0.5cm}

\noindent
\textbf{Task-based Calibration Results.} 
We further use the marker displacement field sequences from the static state and three typical contact events in real-world peg‑insertion tasks \ref{sec: task-based} to fine‑tune the physics parameters in Isaac Lab.
We find that the contact friction coefficient $\mu_{isaac}$, together with the use of compliant contact in Isaac Lab, plays a critical role in reducing the discrepancy between IPC-simulated and real‑world marker displacement fields during contact.
The closest field between reality and simulation for these four task‑based scenarios after calibration are shown in \cref{fig:ipc cali} for these 4 task-based situation after calibration, demonstrating strong alignment.

\subsection{Online RL Learning Results in Simulation}
\begin{figure}[htbp]
  \centering
  \includegraphics[width=1\textwidth]{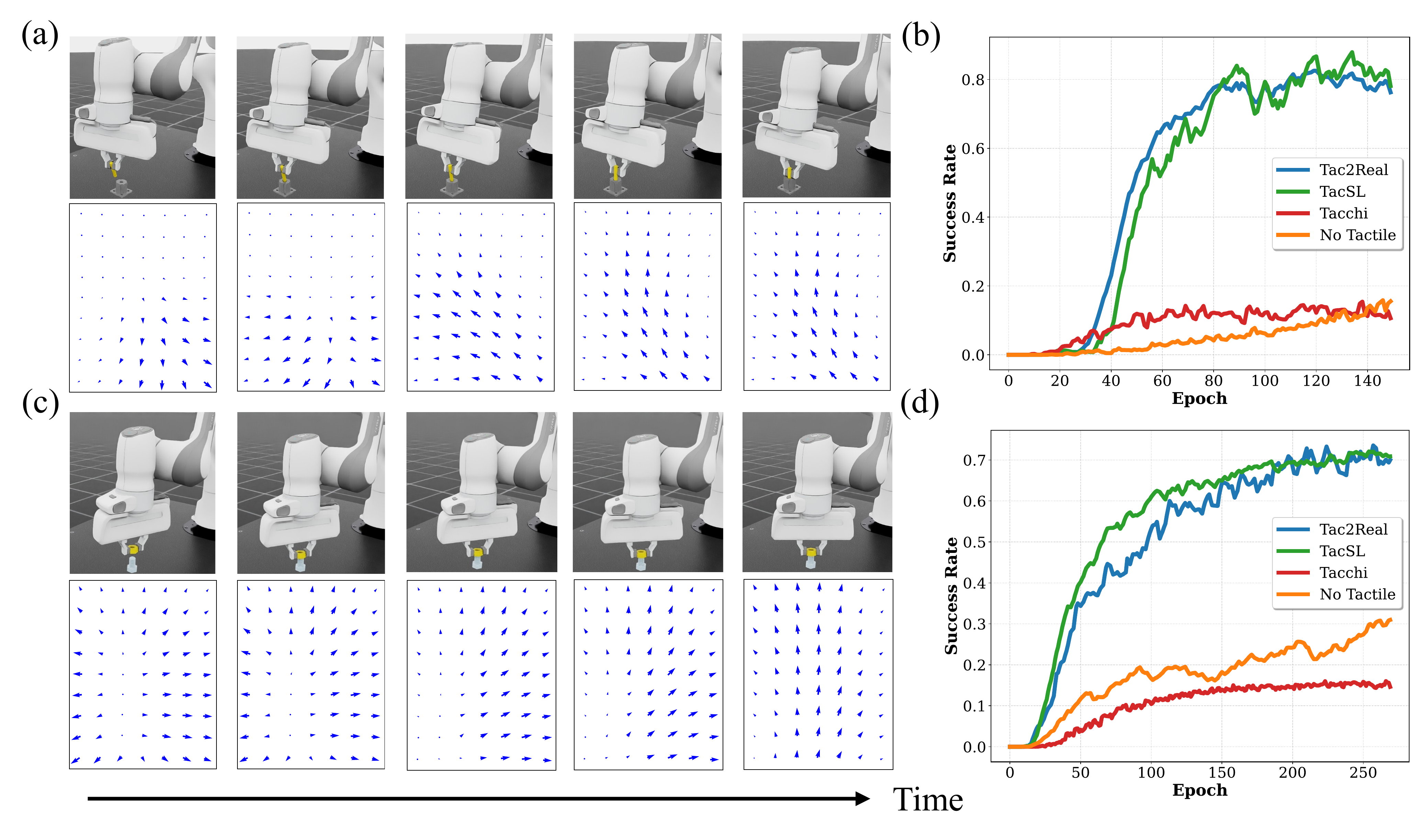}  % 替换为你的 PDF 文件名
  \caption{Contact-rich tasks in simulation. (a)and (c) are snapshots of inference in peg insertion and nut threading tasks with tactile simulation; (b) and (d) are comparisons of the learning curve among Tac2Real, TacSL, Tacchi and no tactile feedback.}
  \label{fig:rl_sim}
\end{figure}

We first evaluate Tac2Real using two typical contact-rich tasks, where tactile marker displacement fields are essential to estimating the relative pose and identifying the complex contact interactions:
\begin{itemize}
    \item[\textbullet] \textbf{Random Orientation Peg Insertion.} 
    The Franka robot is required to fully insert a cylindrical peg held in its gripper into a fixed cylindrical socket.
    Notably, the entire insertion process, especially after the peg enters the socket, must not rely on gravity; that is, the insertion is
    achieved entirely through the robot’s active control rather than gravitational effects, as illustrated in \cref{fig:rl_sim} (a). 
    The initial orientation of the peg in the gripper is randomly sampled from the range of [$-35^{\circ}, 35^{\circ}$], and both the peg and socket hole have a diameter of 8~mm, which is smaller than the real-world setting used in TacSL. 
    \item[\textbullet] \textbf{Random Orientation Nut Threading.} 
    The Franka robot must place a nut (also held with a random initial orientation from $-35^{\circ}$ to $35^{\circ}$) onto a fixed bolt and fasten the nut onto the bolt via gripper rotation, as shown in \cref{fig:rl_sim} (b). 
    Task success is defined as the nut is threaded into the bolt to a depth of 1.5 pitches. 
\end{itemize}

\vspace{-0.5cm}
\vspace{-0.5cm}

The online RL training is formulated as a Markov Decision Process (MDP). To emphasize tactile reliance, the policy observes only the end-effector pose ($\mathbf{p}^{ee} \in \mathbb{R}^7$), the tactile marker displacement field for a single finger ($\mathbf{u} \in \mathbb{R}^{7\times9\times2}$), and the previous action ($\mathbf{a}_{t-1}$), explicitly excluding object pose or visual data. Actions consist of incremental Cartesian updates executed via an impedance controller. The reward function balances task success through keypoint alignment and sparse bonuses for engagement and insertion, while penalizing excessive contact forces to prevent collisions. More training settings are given in supplementary material.

We set the total number of environments to 512, evenly distributing the tactile simulation tasks across four computing nodes, each equipped with 16 GPUs.
We use the off-the-shelf Proximal Policy Optimization (PPO) implementation from the rl-games library in Isaac Lab, where the agent adopts a shared actor–critic architecture.
For a fair comparison, we also integrate Tacchi and TacSL into Isaac Lab using integration schemes analogous to Tac2Real; further details are provided in the supplementary materials.
\cref{fig:rl_sim} shows the average learning curves over three random seeds for the three simulation methods, as well as the setting without tactile observations and the snapshots for the two tasks.
We also report the average success rate over 256 random initial configurations in \cref{tab:rl_sim}.  
Among these, training based on Tac2Real and TacSL exhibit similar performance in simulation environments, both significantly outperforming Tacchi and the tactile-free case.
This demonstrates the effectiveness of our tactile simulation framework and the critical role of tactile sensing in contact-rich tasks, particularly under nearly blind operating conditions.
\vspace{-0.2cm}
\setlength{\tabcolsep}{6pt}      % Increases horizontal space between columns (default is 6pt)
\begin{table}[h]
\centering
\scriptsize   % 选项: \small (稍大), \footnotesize (推荐), \scriptsize (最小)
\caption{Policy Learning Results in both the Simulation and Real-world}
\label{tab:rl_sim}
\begin{tabular}{ccccccc}
\toprule
  &\textbf{Env.}& \textbf{TacAlign level}& \textbf{Tac2Real} & \textbf{TacSL} & \textbf{Tacchi} & \textbf{No Tactile}\\
\midrule
Peg Insertion & sim & - &0.776 & 0.789 & 0.173& 0.168 \\ % [4pt] adds extra space   

Nut Threading & sim & - &0.702& 0.708 & 0.152& 0.313 \\
\midrule
Peg Insertion & real & 1,2,3,4 &\textbf{0.917} & 0.150 & 0.083& 0.067 \\ 
Peg Insertion & real & 2,3,4 &0.533 & 0.033 & 0.050& 0.017 \\
Peg Insertion & real & 1,2,4 &0.250 & 0.150 & 0.016& 0.067 \\
Peg Insertion & real & 1,2,3 &0.767 & 0.100 & 0.100& 0.017 \\

\bottomrule
\end{tabular}
\end{table}
% Reset spacing if needed for subsequent tables (optional if used locally)
% \renewcommand{\arraystretch}{1.0}
\setlength{\tabcolsep}{6pt}
\vspace{-1cm}

\subsection{Real-world Deployment}
\begin{figure}[htbp]
  \centering
  \includegraphics[width=1\textwidth]{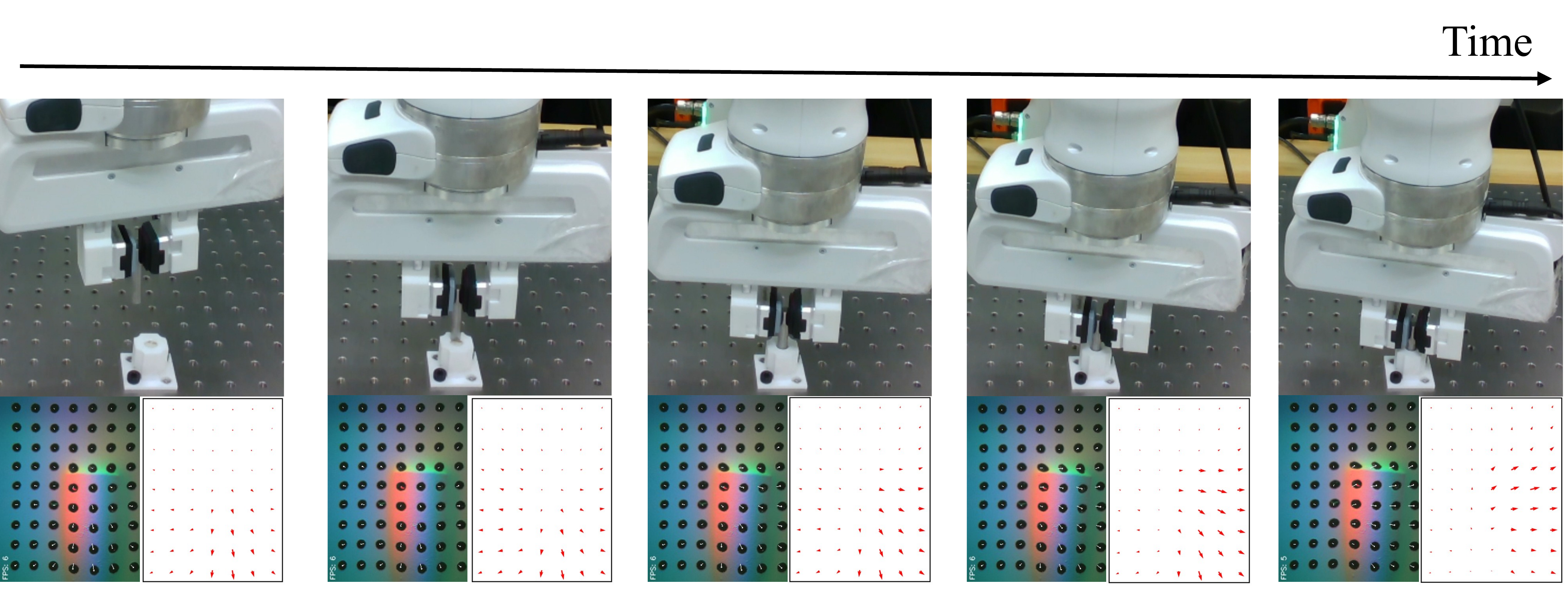}  % 替换为你的 PDF 文件名
  \caption{Peg Insertion Real-world Deployment Snapshots.}
  \label{fig:real_deploy}
\end{figure}

The policy trained in simulation is also deployed in the real world in a zero-shot manner for the peg insertion task.  
We mounted two GelSight Mini sensors on the Franka grippers to maintain force equilibrium, but using only the tactile marker displacement field from the right finger for inference. 
Given the sensor’s gel pad vulnerability to damage, we implement a protective mechanism in practice. 
Specifically, when the mean squared error (MSE) difference between successive marker displacement fields exceeds a predefined threshold, we halt inference, move the end-effector back one step, and resume inference from this adjusted position.
We initialize the peg's initial orientation to $0^{\circ}$ and $\pm15^{\circ}$, and conduct a total of 60 trials with 20 trials for each peg orientation.
As shown in \cref{tab:rl_sim}, we recorded 55 successful insertions, achieving an approximate zero-shot sim-to-real transfer success rate of 91.7\%. 
In \cref{fig:real_deploy}, some inference snapshots of 0-degree situation is illustrated. More snapshots are provided in supplementary material.
Unlike the results in the simulation, the deployment of TacSL only attains a 15\% success rate, attributable to the substantial discrepancy between simulated and real marker displacement fields (see \cref{fig:sim_compare}). 
The Tacchi-based policy acquires low successful insertion counts because of the huge amount of useless tactile feedback caused by MPM numerical instability. 
Additionally, the policy without tactile observations achieved a real-world success rate of only 6.7\%. 
In order to evaluate the importance of TacAlign in sim-to-real transfer, we conducted ablation studies across different levels of TacAlign, where the control alignment, baseline IPC calibration, task-based calibration, and randomization are labeled as levels 1, 2, 3, and 4, respectively. 
The ablation results are presented in \cref{tab:rl_sim}. 
We found that the task-based calibration plays an important role in narrowing the sim-to-real gap, without which the success rate will decrease significantly. 
These results further validate the efficiency of our Tac2Real framework.
\vspace{-0.5cm}
\vspace{-0.5cm}

\section{Conclusion}
In this paper, we present a framework named Tac2Real which contains effctive and robust visuotactile simulation using PNCG-IPC method with  multi-node and multi-GPU highly parallel computing architecture. Tac2Real supports seamless integration with current physics engines to perform online RL training. We also upgrade Tac2Real with a practical and reproducible framework named TacAlign to systematically reduce the sim-to-real gap from structured and stochastic perspectives. Compared with existing methods, Tac2Real achieves robust and high-fidelity tactile simulation, as well as excellent performance in RL policy zero-shot real-world deployment in the peg insertion contact-rich manipulation task. As for future work, Tac2Real can be extended to the RL training for dexterous hands manipulations or deformable objects, and it also can provide large-scale high-quality data for tactile-based vision-language-action (VLA) training. Furthermore, we can consider using AI models to accelerate the tactile simulation to further improve the efficiency.
Finally, the architecture of Tac2real can be generalized to any physical sensor simulation, including tactile, acoustic or even olfactory sensor simulation.

% \newpage
\bibliographystyle{splncs04}
\bibliography{main}

@String(TOG   = {ACM Trans. Graph.})

@String(TOG   = {ACM TOG})

@article{tegin2005tactile,
  title={Tactile sensing in intelligent robotic manipulation--a review},
  author={Tegin, Johan and Wikander, Jan},
  journal={Industrial Robot: An International Journal},
  volume={32},
  number={1},
  pages={64--70},
  year={2005},
  publisher={Emerald Group Publishing Limited}
}

@article{yousef2011tactile,
  title={Tactile sensing for dexterous in-hand manipulation in robotics—A review},
  author={Yousef, Hanna and Boukallel, Mehdi and Althoefer, Kaspar},
  journal={Sensors and Actuators A: physical},
  volume={167},
  number={2},
  pages={171--187},
  year={2011},
  publisher={Elsevier}
}

@article{wang2024multimodal,
  title={Multimodal human--robot interaction for human-centric smart manufacturing: a survey},
  author={Wang, Tian and Zheng, Pai and Li, Shufei and Wang, Lihui},
  journal={Advanced Intelligent Systems},
  volume={6},
  number={3},
  pages={2300359},
  year={2024},
  publisher={Wiley Online Library}
}

@inproceedings{zhao2025polytouch,
  title={Polytouch: A robust multi-modal tactile sensor for contact-rich manipulation using tactile-diffusion policies},
  author={Zhao, Jialiang and Kuppuswamy, Naveen and Feng, Siyuan and Burchfiel, Benjamin and Adelson, Edward},
  booktitle={2025 IEEE International Conference on Robotics and Automation (ICRA)},
  pages={104--110},
  year={2025},
  organization={IEEE}
}

@article{yu2023mimictouch,
  title={Mimictouch: Leveraging multi-modal human tactile demonstrations for contact-rich manipulation},
  author={Yu, Kelin and Han, Yunhai and Wang, Qixian and Saxena, Vaibhav and Xu, Danfei and Zhao, Ye},
  journal={arXiv preprint arXiv:2310.16917},
  year={2023}
}

@article{xue2025reactive,
  title={Reactive diffusion policy: Slow-fast visual-tactile policy learning for contact-rich manipulation},
  author={Xue, Han and Ren, Jieji and Chen, Wendi and Zhang, Gu and Fang, Yuan and Gu, Guoying and Xu, Huazhe and Lu, Cewu},
  journal={arXiv preprint arXiv:2503.02881},
  year={2025}
}

@article{abad2020visuotactile,
  title={Visuotactile sensors with emphasis on gelsight sensor: A review},
  author={Abad, Alexander C and Ranasinghe, Anuradha},
  journal={IEEE Sensors Journal},
  volume={20},
  number={14},
  pages={7628--7638},
  year={2020},
  publisher={IEEE}
}

@inproceedings{yuan2015measurement,
  title={Measurement of shear and slip with a GelSight tactile sensor},
  author={Yuan, Wenzhen and Li, Rui and Srinivasan, Mandayam A and Adelson, Edward H},
  booktitle={2015 IEEE international conference on robotics and automation (ICRA)},
  pages={304--311},
  year={2015},
  organization={IEEE}
}

@inproceedings{yuan2016estimating,
  title={Estimating object hardness with a gelsight touch sensor},
  author={Yuan, Wenzhen and Srinivasan, Mandayam A and Adelson, Edward H},
  booktitle={2016 IEEE/RSJ International Conference on Intelligent Robots and Systems (IROS)},
  pages={208--215},
  year={2016},
  organization={IEEE}
}

@article{sulsky2004implicit,
  title={Implicit dynamics in the material-point method},
  author={Sulsky, Deborah and Kaul, A},
  journal={Computer Methods in Applied Mechanics and Engineering},
  volume={193},
  number={12-14},
  pages={1137--1170},
  year={2004},
  publisher={Elsevier}
}

@article{bardenhagen2004generalized,
  title={The generalized interpolation material point method},
  author={Bardenhagen, Scott G and Kober, Edward M},
  journal={Computer Modeling in Engineering \& Sciences},
  volume={5},
  number={6},
  pages={477},
  year={2004},
  publisher={Tech Science Press}
}

@incollection{jiang2016material,
  title={The material point method for simulating continuum materials},
  author={Jiang, Chenfanfu and Schroeder, Craig and Teran, Joseph and Stomakhin, Alexey and Selle, Andrew},
  booktitle={Acm siggraph 2016 courses},
  pages={1--52},
  year={2016}
}

@article{zhang2025nitsche,
  title={Nitsche-based material point method for large deformation frictional contact problems},
  author={Zhang, Kun and Shen, Shui-Long and Wu, Hui and Zhou, Annan},
  journal={Computational Particle Mechanics},
  volume={12},
  number={2},
  pages={947--970},
  year={2025},
  publisher={Springer}
}

@article{ma2014new,
  title={A new contact algorithm in the material point method for geotechnical simulations},
  author={Ma, J and Wang, Dong and Randolph, MF},
  journal={International Journal for Numerical and Analytical Methods in Geomechanics},
  volume={38},
  number={11},
  pages={1197--1210},
  year={2014},
  publisher={Wiley Online Library}
}

@article{huang2011contact,
  title={Contact algorithms for the material point method in impact and penetration simulation},
  author={Huang, Peng and Zhang, X and Ma, S and Huang, X},
  journal={International journal for numerical methods in engineering},
  volume={85},
  number={4},
  pages={498--517},
  year={2011},
  publisher={Wiley Online Library}
}

@article{wang2022tacto,
  title={Tacto: A fast, flexible, and open-source simulator for high-resolution vision-based tactile sensors},
  author={Wang, Shaoxiong and Lambeta, Mike and Chou, Po-Wei and Calandra, Roberto},
  journal={IEEE Robotics and Automation Letters},
  volume={7},
  number={2},
  pages={3930--3937},
  year={2022},
  publisher={IEEE}
}

@article{de2020material,
  title={Material point method after 25 years: Theory, implementation, and applications},
  author={De Vaucorbeil, Alban and Nguyen, Vinh Phu and Sinaie, Sina and Wu, Jian Ying},
  journal={Advances in applied mechanics},
  volume={53},
  pages={185--398},
  year={2020},
  publisher={Elsevier}
}

@article{lan2021medial,
  title={Medial IPC: accelerated incremental potential contact with medial elastics},
  author={Lan, Lei and Yang, Yin and Kaufman, Danny and Yao, Junfeng and Li, Minchen and Jiang, Chenfanfu},
  journal={ACM Transactions on Graphics},
  volume={40},
  number={4},
  year={2021}
}

@article{chen2023tacchi,
  title={Tacchi: A pluggable and low computational cost elastomer deformation simulator for optical tactile sensors},
  author={Chen, Zixi and Zhang, Shixin and Luo, Shan and Sun, Fuchun and Fang, Bin},
  journal={IEEE Robotics and Automation Letters},
  volume={8},
  number={3},
  pages={1239--1246},
  year={2023},
  publisher={IEEE}
}

@article{sun2025tacchi,
  title={Tacchi 2.0: A low computational cost and comprehensive dynamic contact simulator for vision-based tactile sensors},
  author={Sun, Yuhao and Zhang, Shixin and Li, Wenzhuang and Zhao, Jie and Shan, Jianhua and Shen, Zirong and Chen, Zixi and Sun, Fuchun and Guo, Di and Fang, Bin},
  journal={arXiv preprint arXiv:2503.09100},
  year={2025}
}

@article{du2024tacipc,
  title={Tacipc: Intersection-and inversion-free fem-based elastomer simulation for optical tactile sensors},
  author={Du, Wenxin and Xu, Wenqiang and Ren, Jieji and Yu, Zhenjun and Lu, Cewu},
  journal={IEEE Robotics and Automation Letters},
  volume={9},
  number={3},
  pages={2559--2566},
  year={2024},
  publisher={IEEE}
}

@article{li2020incremental,
  title={Incremental potential contact: intersection-and inversion-free, large-deformation dynamics},
  author={Li, Minchen and Ferguson, Zachary and Schneider, Teseo and Langlois, Timothy and Zorin, Denis and Panozzo, Daniele and Jiang, Chenfanfu and Kaufman, Danny M},
  journal={ACM transactions on graphics},
  year={2020}
}

@article{li2020codimensional,
  title={Codimensional incremental potential contact},
  author={Li, Minchen and Kaufman, Danny M and Jiang, Chenfanfu},
  journal={arXiv preprint arXiv:2012.04457},
  year={2020}
}

@inproceedings{ferguson2023high,
  title={High-order incremental potential contact for elastodynamic simulation on curved meshes},
  author={Ferguson, Zachary and Jain, Pranav and Zorin, Denis and Schneider, Teseo and Panozzo, Daniele},
  booktitle={ACM SIGGRAPH 2023 conference proceedings},
  pages={1--11},
  year={2023}
}

@article{huang2024gipc,
  title={GIPC: Fast and stable Gauss-Newton optimization of IPC barrier energy},
  author={Huang, Kemeng and Chitalu, Floyd M and Lin, Huancheng and Komura, Taku},
  journal={ACM Transactions on Graphics},
  volume={43},
  number={2},
  pages={1--18},
  year={2024},
  publisher={ACM New York, NY}
}

@article{gu2024humanoid,
  title={Humanoid-gym: Reinforcement learning for humanoid robot with zero-shot sim2real transfer},
  author={Gu, Xinyang and Wang, Yen-Jen and Chen, Jianyu},
  journal={arXiv preprint arXiv:2404.05695},
  year={2024}
}

@inproceedings{qureshi2025splatsim,
  title={Splatsim: Zero-shot sim2real transfer of rgb manipulation policies using gaussian splatting},
  author={Qureshi, M Nomaan and Garg, Sparsh and Yandun, Francisco and Held, David and Kantor, George and Silwal, Abhisesh},
  booktitle={2025 IEEE International Conference on Robotics and Automation (ICRA)},
  pages={6502--6509},
  year={2025},
  organization={IEEE}
}

@article{zhao2026high,
  title={High-fidelity simulated data generation for real-world zero-shot robotic manipulation learning with gaussian splatting},
  author={Zhao, Haoyu and Zeng, Cheng and Zhuang, Linghao and Zhao, Yaxi and Xue, Shengke and Wang, Hao and Zhao, Xingyue and Li, Zhongyu and Li, Kehan and Huang, Siteng and others},
  journal={IEEE Robotics and Automation Letters},
  year={2026},
  publisher={IEEE}
}

@article{zhong2024tactgen,
  title={TactGen: tactile sensory data generation via zero-shot sim-to-real transfer},
  author={Zhong, Shaohong and Albini, Alessandro and Maiolino, Perla and Posner, Ingmar},
  journal={IEEE Transactions on Robotics},
  volume={41},
  pages={1316--1328},
  year={2024},
  publisher={IEEE}
}

@article{si2024difftactile,
  title={Difftactile: A physics-based differentiable tactile simulator for contact-rich robotic manipulation},
  author={Si, Zilin and Zhang, Gu and Ben, Qingwei and Romero, Branden and Xian, Zhou and Liu, Chao and Gan, Chuang},
  journal={arXiv preprint arXiv:2403.08716},
  year={2024}
}

@book{zienkiewicz1977finite,
  title={The finite element method},
  author={Zienkiewicz, Olgierd Cecil and Taylor, Robert Leroy and Nithiarasu, Perumal and Zhu, JZ},
  volume={3},
  year={1977},
  publisher={Elsevier}
}

@article{sun2025soft,
  title={Soft contact simulation and manipulation learning of deformable objects with vision-based tactile sensor},
  author={Sun, Yuhao and Zhang, Shixin and Chen, Zixi and Shen, Zirong and Sun, Fuchun and Stefanini, Cesare and Guo, Di and Luo, Shan and Zhang, Jianwei and Shan, Jianhua and others},
  journal={IEEE Transactions on Automation Science and Engineering},
  year={2025},
  publisher={IEEE}
}

@article{chen2024general,
  title={General-purpose sim2real protocol for learning contact-rich manipulation with marker-based visuotactile sensors},
  author={Chen, Weihang and Xu, Jing and Xiang, Fanbo and Yuan, Xiaodi and Su, Hao and Chen, Rui},
  journal={IEEE Transactions on Robotics},
  volume={40},
  pages={1509--1526},
  year={2024},
  publisher={IEEE}
}

@article{akinola2025tacsl,
  title={Tacsl: A library for visuotactile sensor simulation and learning},
  author={Akinola, Iretiayo and Xu, Jie and Carius, Jan and Fox, Dieter and Narang, Yashraj},
  journal={IEEE Transactions on Robotics},
  year={2025},
  publisher={IEEE}
}

@article{makoviychuk2021isaac,
  title={Isaac gym: High performance gpu-based physics simulation for robot learning},
  author={Makoviychuk, Viktor and Wawrzyniak, Lukasz and Guo, Yunrong and Lu, Michelle and Storey, Kier and Macklin, Miles and Hoeller, David and Rudin, Nikita and Allshire, Arthur and Handa, Ankur and others},
  journal={arXiv preprint arXiv:2108.10470},
  year={2021}
}

@article{zhang2025tacflex,
  title={Tacflex: Multi-mode tactile imprints simulation for visuotactile sensors with coating patterns},
  author={Zhang, Chaofan and Cui, Shaowei and Hu, Jingyi and Jiang, Tianyu and Zhang, Tiandong and Wang, Rui and Wang, Shuo},
  journal={IEEE Transactions on Robotics},
  year={2025},
  publisher={IEEE}
}

@article{xu2021end,
  title={An end-to-end differentiable framework for contact-aware robot design},
  author={Xu, Jie and Chen, Tao and Zlokapa, Lara and Foshey, Michael and Matusik, Wojciech and Sueda, Shinjiro and Agrawal, Pulkit},
  journal={arXiv preprint arXiv:2107.07501},
  year={2021}
}

@inproceedings{xu2023efficient,
  title={Efficient tactile simulation with differentiability for robotic manipulation},
  author={Xu, Jie and Kim, Sangwoon and Chen, Tao and Garcia, Alberto Rodriguez and Agrawal, Pulkit and Matusik, Wojciech and Sueda, Shinjiro},
  booktitle={Conference on Robot Learning},
  pages={1488--1498},
  year={2023},
  organization={PMLR}
}

@article{li2025taccel,
  title={Taccel: Scaling up vision-based tactile robotics via high-performance gpu simulation},
  author={Li, Yuyang and Du, Wenxin and Yu, Chang and Li, Puhao and Zhao, Zihang and Liu, Tengyu and Jiang, Chenfanfu and Zhu, Yixin and Huang, Siyuan},
  journal={arXiv preprint arXiv:2504.12908},
  year={2025}
}

@article{dai2013nonlinear,
  title={A nonlinear conjugate gradient algorithm with an optimal property and an improved Wolfe line search},
  author={Dai, Yu-Hong and Kou, Cai-Xia},
  journal={SIAM Journal on Optimization},
  volume={23},
  number={1},
  pages={296--320},
  year={2013},
  publisher={SIAM}
}

@book{dhatt2012finite,
  title={Finite element method},
  author={Dhatt, Gouri and Lefran{\c{c}}ois, Emmanuel and Touzot, Gilbert},
  year={2012},
  publisher={John Wiley \& Sons}
}

@article{hu2019taichi,
  title={Taichi: a language for high-performance computation on spatially sparse data structures},
  author={Hu, Yuanming and Li, Tzu-Mao and Anderson, Luke and Ragan-Kelley, Jonathan and Durand, Fr{\'e}do},
  journal={ACM Transactions on Graphics (TOG)},
  volume={38},
  number={6},
  pages={1--16},
  year={2019},
  publisher={ACM New York, NY, USA}
}

@inproceedings{kim2022ipc,
  title={Ipc-graspsim: Reducing the sim2real gap for parallel-jaw grasping with the incremental potential contact model},
  author={Kim, Chung Min and Danielczuk, Michael and Huang, Isabella and Goldberg, Ken},
  booktitle={2022 International Conference on Robotics and Automation (ICRA)},
  pages={6180--6187},
  year={2022},
  organization={IEEE}
}

@article{zhang2020modular,
title={A modular robotic arm control stack for research: Franka-interface and frankapy},
author={Zhang, Kevin and Sharma, Mohit and Liang, Jacky and Kroemer, Oliver},
journal={arXiv preprint arXiv:2011.02398},
year={2020}
}

@inproceedings{shen2024preconditioned,
  title={Preconditioned nonlinear conjugate gradient method for real-time interior-point hyperelasticity},
  author={Shen, Xing and Cai, Runyuan and Bi, Mengxiao and Lv, Tangjie},
  booktitle={ACM SIGGRAPH 2024 Conference Papers},
  pages={1--11},
  year={2024}
}

@article{mittal2025isaaclab,
  title={Isaac Lab: A GPU-Accelerated Simulation Framework for Multi-Modal Robot Learning},
  author={Mayank Mittal and Pascal Roth and James Tigue and Antoine Richard and Octi Zhang and Peter Du and Antonio Serrano-Muñoz and Xinjie Yao and René Zurbrügg and Nikita Rudin and Lukasz Wawrzyniak and Milad Rakhsha and Alain Denzler and Eric Heiden and Ales Borovicka and Ossama Ahmed and Iretiayo Akinola and Abrar Anwar and Mark T. Carlson and Ji Yuan Feng and Animesh Garg and Renato Gasoto and Lionel Gulich and Yijie Guo and M. Gussert and Alex Hansen and Mihir Kulkarni and Chenran Li and Wei Liu and Viktor Makoviychuk and Grzegorz Malczyk and Hammad Mazhar and Masoud Moghani and Adithyavairavan Murali and Michael Noseworthy and Alexander Poddubny and Nathan Ratliff and Welf Rehberg and Clemens Schwarke and Ritvik Singh and James Latham Smith and Bingjie Tang and Ruchik Thaker and Matthew Trepte and Karl Van Wyk and Fangzhou Yu and Alex Millane and Vikram Ramasamy and Remo Steiner and Sangeeta Subramanian and Clemens Volk and CY Chen and Neel Jawale and Ashwin Varghese Kuruttukulam and Michael A. Lin and Ajay Mandlekar and Karsten Patzwaldt and John Welsh and Huihua Zhao and Fatima Anes and Jean-Francois Lafleche and Nicolas Moënne-Loccoz and Soowan Park and Rob Stepinski and Dirk Van Gelder and Chris Amevor and Jan Carius and Jumyung Chang and Anka He Chen and Pablo de Heras Ciechomski and Gilles Daviet and Mohammad Mohajerani and Julia von Muralt and Viktor Reutskyy and Michael Sauter and Simon Schirm and Eric L. Shi and Pierre Terdiman and Kenny Vilella and Tobias Widmer and Gordon Yeoman and Tiffany Chen and Sergey Grizan and Cathy Li and Lotus Li and Connor Smith and Rafael Wiltz and Kostas Alexis and Yan Chang and David Chu and Linxi "Jim" Fan and Farbod Farshidian and Ankur Handa and Spencer Huang and Marco Hutter and Yashraj Narang and Soha Pouya and Shiwei Sheng and Yuke Zhu and Miles Macklin and Adam Moravanszky and Philipp Reist and Yunrong Guo and David Hoeller and Gavriel State},
  journal={arXiv preprint arXiv:2511.04831},
  year={2025},
  url={https://arxiv.org/abs/2511.04831}
}

@inproceedings{todorov2012mujoco,
  title={MuJoCo: A physics engine for model-based control},
  author={Todorov, Emanuel and Erez, Tom and Tassa, Yuval},
  booktitle={2012 IEEE/RSJ International Conference on Intelligent Robots and Systems},
  pages={5026--5033},
  year={2012},
  organization={IEEE},
  doi={10.1109/IROS.2012.6386109}
}

@article{hietala2021closing,
  title={Closing the sim2real gap in dynamic cloth manipulation},
  author={Hietala, Julius and Blanco-Mulero, David and Alcan, Gokhan and Kyrki, Ville},
  journal={arXiv preprint arXiv:2109.04771},
  volume={2},
  year={2021}
}

@article{zhao2026closing,
  title={Closing the Reality Gap: Zero-Shot Sim-to-Real Deployment for Dexterous Force-Based Grasping and Manipulation},
  author={Zhao, Zhe and Dong, Haoyu and He, Zhengmao and Li, Yang and Yi, Xinyu and Li, Zhibin},
  journal={arXiv preprint arXiv:2601.02778},
  year={2026}
}

@article{blanco2024benchmarking,
  title={Benchmarking the sim-to-real gap in cloth manipulation},
  author={Blanco-Mulero, David and Barbany, Oriol and Alcan, Gokhan and Colom{\'e}, Adri{\`a} and Torras, Carme and Kyrki, Ville},
  journal={IEEE Robotics and Automation Letters},
  volume={9},
  number={3},
  pages={2981--2988},
  year={2024},
  publisher={IEEE}
}

@inproceedings{josifovski2022analysis,
  title={Analysis of randomization effects on sim2real transfer in reinforcement learning for robotic manipulation tasks},
  author={Josifovski, Josip and Malmir, Mohammadhossein and Klarmann, Noah and {\v{Z}}agar, Bare Luka and Navarro-Guerrero, Nicol{\'a}s and Knoll, Alois},
  booktitle={2022 IEEE/RSJ International Conference on Intelligent Robots and Systems (IROS)},
  pages={10193--10200},
  year={2022},
  organization={IEEE}
}

@misc{rl-games2021,
title = {rl-games: A High-performance Framework for Reinforcement Learning},
author = {Makoviichuk, Denys and Makoviychuk, Viktor},
month = {May},
year = {2021},
publisher = {GitHub},
journal = {GitHub repository},
howpublished = {\url{https://github.com/Denys88/rl_games}},
}

@inproceedings{moritz2018ray,
  title={Ray: A distributed framework for emerging $\{$AI$\}$ applications},
  author={Moritz, Philipp and Nishihara, Robert and Wang, Stephanie and Tumanov, Alexey and Liaw, Richard and Liang, Eric and Elibol, Melih and Yang, Zongheng and Paul, William and Jordan, Michael I and others},
  booktitle={13th USENIX symposium on operating systems design and implementation (OSDI 18)},
  pages={561--577},
  year={2018}
}

@article{hansen2006cma,
  title={The CMA evolution strategy: a comparing review},
  author={Hansen, Nikolaus},
  journal={Towards a new evolutionary computation: Advances in the estimation of distribution algorithms},
  pages={75--102},
  year={2006},
  publisher={Springer}
}

\newpage
\begin{center}
    \Large\bfseries Supplementary Materials
\end{center}
\setcounter{section}{0} % 重置计数器为0（下一个就是1->A）
\renewcommand{\thesection}{\Alph{section}}
\section{Detailed Introduction to PNGC-IPC}
\label{sec:intro_PNCG_IPC}

Our tactile simulation is built upon the PNCG-IPC solver~\cite{shen2024preconditioned}, which achieves a balance between physical accuracy and computational efficiency for simulating deformable bodies in contact-rich scenarios.
The simulation of the elastomer in visuotactile sensors requires modeling a hyperelastic body undergoing large deformations with frictional contact. 
We discretize the sensor gel using a tetrahedral mesh and employ implicit Euler integration, which formulates each time step as a constrained optimization: minimizing the sum of inertial and elastic potential energies subject to collision-free constraints $h_l(\mathbf{x})\geq 0$ between all surface primitives:
\begin{equation}
  \mathbf{x}^{t+1} = \arg\min_{\mathbf{x}} \; \frac{1}{2}(\mathbf{x}-\hat{\mathbf{x}})^\top \mathbf{M}(\mathbf{x}-\hat{\mathbf{x}}) + h^2\Psi(\mathbf{x}) \quad \text{s.t.} \;\; h_l(\mathbf{x}) \geq 0,
  \label{eq:varopt}
\end{equation}
where $\mathbf{M}$ is the mass matrix, $h$ the time step size, $\hat{\mathbf{x}} = \mathbf{x}^t + h\mathbf{v}^t$ the inertial prediction, and $\Psi(\mathbf{x})$ the hyperelastic potential.
IPC~\cite{li2020incremental, li2020codimensional, ferguson2023high, kim2022ipc} relaxes the hard constraints into an unconstrained problem by augmenting the objective with a smooth log-barrier contact potential:
\begin{equation}
  E(\mathbf{x}) = \frac{1}{2}(\mathbf{x}-\hat{\mathbf{x}})^\top \mathbf{M}(\mathbf{x}-\hat{\mathbf{x}}) + h^2\Psi(\mathbf{x}) + \kappa \sum_{k\in C} b\!\left(d_k(\mathbf{x})\right) + D(\mathbf{x}),
  \label{eq:ipc_energy}
\end{equation}
where $d_k$ is the distance between a contact primitive pair, $\kappa$ the barrier stiffness, and $b$ a $C^2$ log-barrier that activates when $d_k$ falls below the threshold $\hat{d}$.
The last term $D(\mathbf{x})$ is the friction dissipation potential that models Coulomb frictional contact. Following~\cite{li2020incremental}, $D$ is defined as:
\begin{equation}
  D(\mathbf{x}) = \mu_f \sum_{k\in C} \lambda_k^n \, f\!\left(\left\|\mathbf{T}_k \Delta \mathbf{x}_k\right\|\right),
  \label{eq:friction_energy}
\end{equation}
where $\mu_f$ is the coefficient of friction, $\lambda_k^n = -\kappa\, b'(d_k)$ is the normal contact force magnitude derived from the barrier gradient, $\mathbf{T}_k$ is the tangent-plane projection operator at contact pair $k$, $\Delta \mathbf{x}_k = \mathbf{x}_k - \mathbf{x}_k^t$ is the relative tangential displacement within the current time step, and $f(\cdot)$ is a $C^1$ smooth mollifier that approximates the non-smooth Coulomb friction cone:
\begin{equation}
  f(s) = \begin{cases} -\frac{s^3}{3\epsilon_v^2} + \frac{s^2}{\epsilon_v} + \frac{\epsilon_v}{3}, & s < \epsilon_v, \\ s, & s \geq \epsilon_v, \end{cases}
  \label{eq:friction_mollifier}
\end{equation}
with $\epsilon_v$ being a small velocity threshold that ensures differentiability near zero sliding velocity. This smooth approximation enables gradient-based optimization while faithfully capturing static and dynamic frictional behavior.
The Hessian of $E$ is $\mathbf{H} = \mathbf{M} + h^2 \nabla^2\Psi + \kappa \sum_{k\in C} \nabla^2 b + \nabla^2 D$.
Standard IPC solves this with Newton's method and continuous collision detection (CCD)-based line search, which is accurate but computationally expensive and difficult to parallelize on GPUs.

PNCG-IPC~\cite{shen2024preconditioned} replaces Newton's method with the Dai-Kou (DK) nonlinear conjugate gradient algorithm~\cite{dai2013nonlinear}, directly minimizing \cref{eq:ipc_energy} without assembling or factorizing $\mathbf{H}$. 
The preconditioned search direction and DK conjugate parameter are:
\begin{equation}
  \mathbf{p}_{k+1} = -\mathbf{P}_{k+1}\mathbf{g}_{k+1} + \beta_k^{DK}\mathbf{p}_k, \quad
  \beta_k^{DK} = \frac{\mathbf{g}_{k+1}^\top \mathbf{P}_{k+1} \mathbf{y}_k}{\mathbf{y}_k^\top  \mathbf{p}_k} - \frac{\mathbf{y}_k^\top \mathbf{P}_{k+1} \mathbf{y}_k}{\mathbf{y}_k^\top \mathbf{p}_k}\frac{\mathbf{p}_k^\top \mathbf{g}_{k+1}}{\mathbf{y}_k^\top \mathbf{p}_k},
  \label{eq:dk_direction}
\end{equation}
where $\mathbf{g}_k$ is the gradient, $\mathbf{y}_k = \mathbf{g}_{k+1} - \mathbf{g}_k$, and $\mathbf{P} = \mathrm{diag}(\mathbf{H})^{-1}$ is the Jacobi preconditioner. 
The hyperelastic Hessian is decomposed into six independent terms per element, so both $\mathrm{diag}(\mathbf{H})$ and $\mathbf{p}^\top\mathbf{H}\mathbf{p}$ can be computed in parallel with minimal FLOPs. 
For the step size, an analytical upper bound $\alpha_{\mathrm{upper}} = \hat{d}/(2\|\mathbf{p}_{k+1}\|_\infty)$ eliminates costly CCD queries, and the optimal size is obtained via:
\begin{equation}
  \bar{\alpha} = -\frac{\mathbf{g}_{k+1}^\top \mathbf{p}_{k+1}}{\mathbf{p}_{k+1}^\top \mathbf{H}_{k+1} \mathbf{p}_{k+1}}, \quad \alpha = \min(\alpha_{\mathrm{upper}},\; \bar{\alpha}).
  \label{eq:step_size}
\end{equation}

This design trades per-iteration accuracy for substantially higher throughput: each conjugate gradient step is less precise than a Newton step, yet the low per-iteration cost on GPUs enables convergence to sufficient accuracy within tens of iterations—a favorable trade-off for tactile simulation where physically consistent deformation matters more than machine-precision convergence.
PNCG-IPC is implemented as a lightweight Taichi~\cite{hu2019taichi}-based solver that compiles Python code to high-performance GPU kernels, making it straightforward to integrate as a plug-in module into diverse simulation pipelines.

\section{Results for PNCG-IPC Simulation}
\begin{figure}[htbp]
  \centering
  \includegraphics[width=1\textwidth]{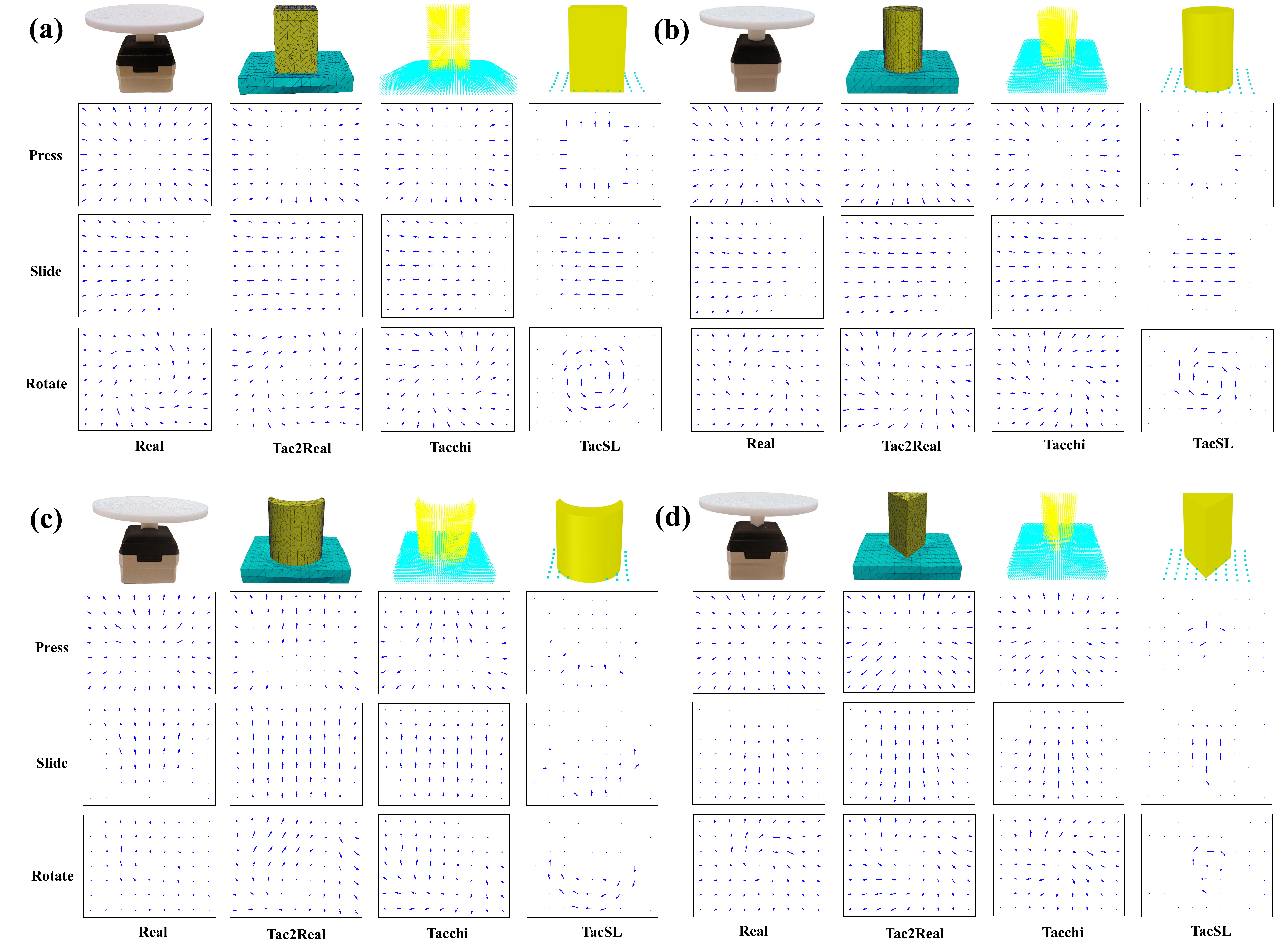}  % 替换为你的 PDF 文件名
  \caption{Supplementary results for PNCG-IPC simulations under 3 deformation modes and 4 indenters. (a) Cube indenter results; (b) Cylinder indenter results; (c) Moon-shape indentert results; (d) Triangle-shape results.}
  \label{fig:s1}
\end{figure}

\begin{figure}[htbp]
  \centering
  \includegraphics[width=1.0\textwidth]{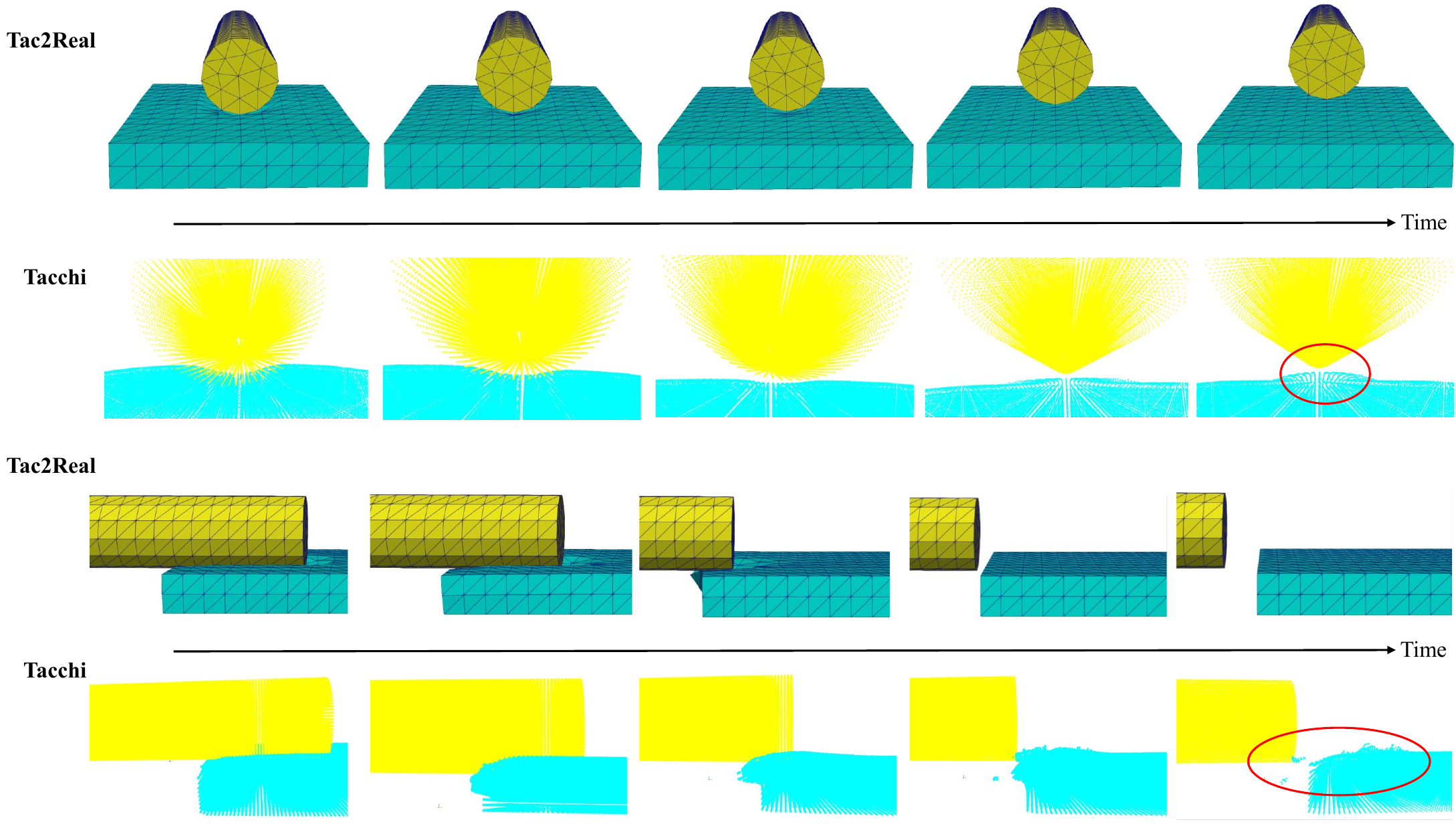}  % 替换为你的 PDF 文件名
  \caption{Comparison of Tac2Real and Tacchi in upward (upper part) and slide (lower part) deformations.}
  \label{fig:s2}
\end{figure}

In this section, we illustrate the tactile simulation results for all the 4 indenters (cube, cylinder, moon and triangle), and compare the our results with real, Tacchi \cite{chen2023tacchi} and TacSL \cite{akinola2025tacsl}. As shown in \cref{fig:s1}, Tac2Real and Tacchi have similar performance under such small press, slide and rotate deformation, while non-physics TacSL presents relative large gap to real-world. We also demonstrate detailed comparison between Tac2Real and Tacchi under upward and large slide deformation, shown in \cref{fig:s2}. In both two cases, Tac2Real shows strong recovery property in the absence of object contact, which approaches to the real-world. However, Tacchi suffers from an adhesion issue when two objects are no longer in contact or even particles splashing. These results verify the reliability and stability of Tac2Real simulation.

\section{Results for Parameters Calibration}
\begin{figure}[htbp]
  \centering
  \includegraphics[width=0.8\textwidth]{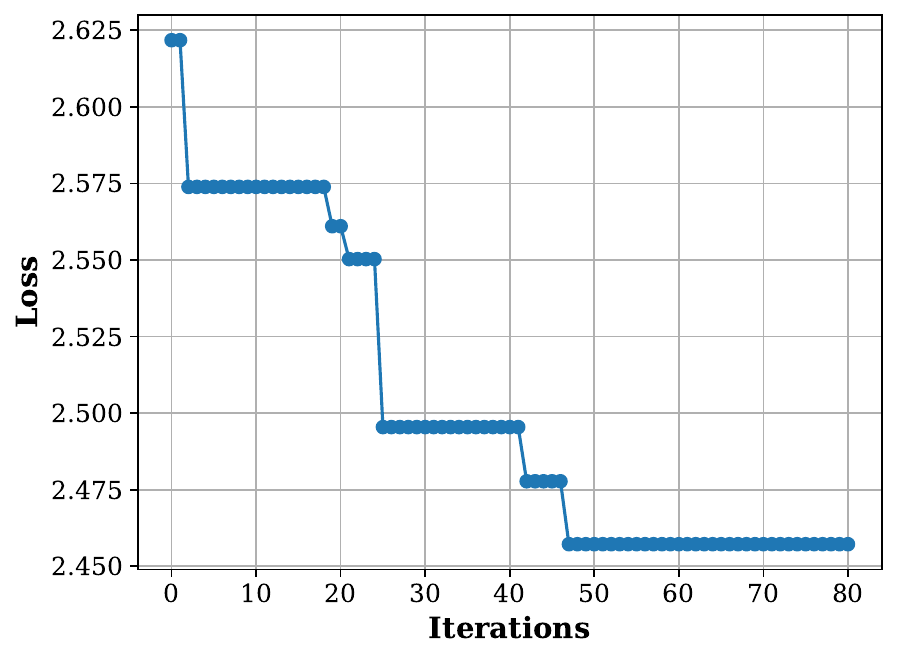}  % 替换为你的 PDF 文件名
  \caption{CMA-ES Loss.}
  \label{fig:cma_es}
\end{figure}

As for the baseline alignment in TacAlign, we choose the range of PNCG-IPC simulation parameters as in \cref{tab:params_range}. We found the 4 parameters differ by orders of magnitude, which are all normalized to [0, 1] in the CMA-ES optimization process to improve searching efficiency. The popsize is set to 12, and he loss curve of CMA-ES after 80 iterations is shown in \cref{fig:cma_es}. The magnitude of the loss reduction is nearly consistent with the CMA-ES in the work \cite{si2024difftactile}.
\setlength{\tabcolsep}{8pt}      % Increases horizontal space between columns (default is 6pt)
\begin{table}[h]
\centering
\footnotesize   % 选项: \small (稍大), \footnotesize (推荐), \scriptsize (最小)
\caption{Parameters Range}
\label{tab:params_range}
\begin{tabular}{ccc}
\toprule
\textbf{Parameter} & \textbf{Lower Bound} & \textbf{Upper Bound} \\
\midrule
$E$ &  1e4 & 2e5 \\ % [4pt] adds extra space after this specific row
$\nu$ & 0.4 & 0.497 \\
$\rho$ & 1e-3 & 5e-3 \\
$\mu$ & 0.25 & 2.5 \\
\bottomrule
\end{tabular}
\end{table}

\section{Integration Details of Tacchi and TacSL}
Both Tacchi and TacSL integrations to Isaac Lab follow the whole pipeline illustrated in \cref{fig:sim frame} by substituting the simulation method with MPM and the penalty-based method, respectively.

% \textbf{Tacchi Integration}. Tacchi \cite{chen2023tacchi} uses MPM implemented on Taichi \cite{hu2019taichi} to perform tactile simulation. All simulation environments in a GPU are embedded in a large grid. Considering the particle-to-grid (P2G) and grid-to-particle (G2P) process in MPM, we firstly need to cut the range of holding objects' particles in order to ensure that each environment does not affect the others. As for the boundary condition, we make the velocity of the lowest grid layer 0 to improve simulation stability, which will fix the lower part of the sensor elastomer, rather than directly limiting the particles' velocity. In online RL training, Tacchi often suffers from numerical explosion. In order to maintain the training process, we check the particle velocity in each step. If the maximal particle velocity exceeds a large threshold, we set all particles' velocities in this environment to 0, ortherwise the training process will be interrupted by simulation collapse.   
\subsection{Tacchi Integration.}
Tacchi~\cite{chen2023tacchi} adopts the explicit MPM,
implemented in Taichi~\cite{hu2019taichi}, to simulate elastomer deformation.
For completeness, we briefly introduce the core MPM procedure used in Tacchi.

MPM discretizes the deformable body into a set of material particles carrying
mass, velocity, and deformation states, while momentum exchange is performed
through a background Eulerian grid. At each time step, the algorithm consists of
three standard stages: particle-to-grid (P2G), grid update, and grid-to-particle
(G2P).

In the P2G stage, particle quantities are transferred to nearby grid nodes:
\begin{equation}
m_i = \sum_p w_{ip} m_p,
\qquad
(m\mathbf{v})_i = \sum_p w_{ip} m_p \mathbf{v}_p ,
\end{equation}
where $m_p$ and $\mathbf{v}_p$ are the mass and velocity of particle $p$,
$m_i$ and $\mathbf{v}_i$ denote the mass and velocity on grid node $i$, and
$w_{ip}$ is the interpolation weight between particle $p$ and node $i$.

The internal elastic force can be assembled on the grid from the particle stress:
\begin{equation}
\mathbf{f}^{\mathrm{int}}_i
=
-
\sum_p
V_p \,
\boldsymbol{\sigma}_p
\nabla w_{ip},
\end{equation}
where $V_p$ is the particle volume and $\boldsymbol{\sigma}_p$ is the Cauchy
stress of particle $p$. Then the grid velocity is updated by
\begin{equation}
\mathbf{v}_i^{\,n+1}
=
\mathbf{v}_i^{\,n}
+
\Delta t \,
\frac{\mathbf{f}^{\mathrm{int}}_i + \mathbf{f}^{\mathrm{ext}}_i}{m_i}.
\end{equation}

In the G2P stage, the updated grid quantities are interpolated back to particles:
\begin{equation}
\mathbf{v}_p^{\,n+1}
=
\sum_i w_{ip}\mathbf{v}_i^{\,n+1},
\qquad
\mathbf{x}_p^{\,n+1}
=
\mathbf{x}_p^{\,n}
+
\Delta t
\sum_i w_{ip}\mathbf{v}_i^{\,n+1}.
\end{equation}
Meanwhile, the deformation gradient is updated as
\begin{equation}
\mathbf{F}_p^{\,n+1}
=
\left(
\mathbf{I}
+
\Delta t \sum_i \mathbf{v}_i^{\,n+1} (\nabla w_{ip})^\top
\right)
\mathbf{F}_p^{\,n}.
\end{equation}

These equations highlight the main characteristic of MPM: the deformable body
is represented by particles, while force computation and momentum update are
performed on the background grid. This hybrid particle-grid formulation makes
MPM effective for large-deformation simulation, but in contact-rich tactile
scenarios it may also suffer from numerical instability, especially under strong
shear, large rotations, or repeated contact.

For integration into Isaac Lab, all simulation environments assigned to one GPU
are embedded into a single large background grid. Considering the P2G and G2P
processes in MPM, we first restrict the spatial range of object particles for each
environment to prevent cross-environment interference. As boundary condition,
we set the velocity of the lowest grid layer to zero, which effectively fixes the
bottom part of the sensor elastomer and improves numerical stability. This
treatment is more stable than directly clamping particle velocities.

During online RL training, Tacchi may still occasionally encounter numerical
explosion. To prevent the entire training process from collapsing, we monitor the
particle velocities at every step. If the maximal particle velocity in one
environment exceeds a prescribed threshold, all particle velocities in that
environment are reset to zero before continuing simulation.

\subsection{TacSL Integration}
% A key part of the TacSL simulation is the signed distance field (SDF) query. The original implementation of TacSL is highly integrated with Isaac Gym, and use its built-in GPU parallel SDF query function. In our work, the non-physics penalty-based simulation is separated from TacSL, including the calculation of tactile force field from penetration distance. We use Taichi to perform the parallel SDF query kernel function to accelerate the RL training. Specifically, the holding object is fixed, and the tactile markers from all environments are transformed according to the relative pose to perform the SDF query simultaneously.
TacSL~\cite{akinola2025tacsl} does not explicitly solve a high-fidelity
continuum deformation field of the elastomer as in IPC- or MPM-based
simulators. Instead, it adopts a fast penalty-based approximation built on top
of a compliant rigid-body contact model. In TacSL, the sensor membrane and the
indenting object are both treated as rigid bodies, while strict non-penetration
is relaxed by a soft contact law so that a small amount of interpenetration is
allowed. The underlying normal contact force in the physics solver follows a
Kelvin--Voigt model,
\begin{equation}
f = \max(-\kappa \epsilon - c \dot{\epsilon},\, 0),
\end{equation}
where $\epsilon$ and $\dot{\epsilon}$ denote the contact distance and separation
velocity, and $\kappa$ and $c$ are the stiffness and damping coefficients.

For tactile signal extraction, TacSL samples a set of tactile points on the
sensor surface and computes a normal/shear contact response at each point using
a penalty-based tactile model. Specifically, the normal and tangential forces
are computed as
\begin{equation}
\mathbf{f}_n = (-k_n d + k_d \dot{d}) \mathbf{n},
\qquad
\mathbf{f}_t =
-\frac{\mathbf{v}_t}{\|\mathbf{v}_t\|}
\min\!\left(k_t \|\mathbf{v}_t\|,\; \mu \|\mathbf{f}_n\|\right),
\end{equation}
where $d$ is the interpenetration distance, $\dot{d}$ is the interpenetration
velocity, $\mathbf{n}$ is the contact normal, $\mathbf{v}_t$ is the tangential
relative velocity, $k_n$ and $k_d$ are the normal stiffness and damping, and
$k_t$ and $\mu$ are the tangential stiffness and friction coefficient,
respectively.

A key component in this computation is the signed distance field (SDF) of the
contacting object. Prior to simulation, TacSL precomputes both the object SDF
and its gradient. At each simulation step, the interpenetration distance at a
tactile point is obtained by querying the SDF,
\begin{equation}
d = \mathrm{SDF}(\mathbf{x}),
\end{equation}
and the normal direction is approximated from the SDF gradient,
\begin{equation}
\mathbf{n} = \nabla d.
\end{equation}
The interpenetration velocity is then computed through the chain rule as
\begin{equation}
\dot{d} = (\nabla d)^{\top}\dot{\mathbf{x}},
\end{equation}
where $\dot{\mathbf{x}}$ is the relative velocity at the queried tactile point.
Therefore, TacSL does not evolve a dense displacement field inside the
elastomer; rather, it estimates local penetration-based tactile responses on
sampled surface points and interprets them as a tactile force field. 
Considering the high similarity and normalization of the force field and displacement field for hyperelastic bodies, we approximate the tangential force field of TacSL as the tangential displacement field.

In our implementation, the original TacSL module is decoupled from Isaac Gym
and re-implemented as a lightweight tactile backend compatible with Isaac Lab.
Since the original method relies heavily on GPU-parallel SDF queries, we
preserve this design and implement the SDF lookup as a Taichi kernel. During
simulation, the grasped object is kept fixed in its local frame, while the
tactile points from all parallel environments are transformed by the current
relative pose between the object and the sensor. Their SDF values and gradients
are then queried simultaneously on the GPU, after which the penalty-based model
is applied to compute the tactile normal/shear response for each point.

\section{Detailed Online RL Settings}
We formulate peg-in-hole as an Markov decision process (MDP).
The state space contains the entire state of the robot and environment, such as robot configuration, object poses, and tactile displacements, while the policy only receives partial observations. 

\setlength{\tabcolsep}{8pt}      % Increases horizontal space between columns (default is 6pt)
\begin{table}[h]
\centering
\scriptsize   % 选项: \small (稍大), \footnotesize (推荐), \scriptsize (最小)
\caption{Parameters Randomization Range}
\label{tab:random}
\begin{tabular}{cc}
\toprule
\textbf{Parameter} & \textbf{Randomization Range}  \\
\midrule
Controller gains $k_p$ &  [400, 800]  \\ % [4pt] adds extra space after this specific row
Peg Friction & [0.5, 1.0] \\
Socket Init. Position (cm) & X:[59.75, 69.25], Y:[-0.25, 0.25], Z:[4.75, 5.25] \\
Holding Init. Position (cm)   & X:[-0.3, 0.3], Z:[-0.3, 0.3] \\
Holding Init. Orientation Y-rot (degree) & [-35, 35] \\
Franka Hand Init. Position (cm) & X:[-2, 2], Y:[-2, 2], Z:[6.5, 8.5] \\
Franka Hand Init. Orientation (rad) & X:3.1415, Y:0.0, Z:[-0.785, 0.785]  \\
\midrule
EE Pose Noise in Translation (mm) & 5 \\
EE Pose Noise in Rotation (rad) & 0.2 \\
IPC Rand. Move. Noise in Translation & 1 \\
IPC Rand. Move. Noise in Rotation & 0.05 \\
\bottomrule
\end{tabular}
\end{table}

\noindent
\textbf{Observation.}
To highlight the role of tactile sensing, the policy observes only:
(i) end-effector pose $p^{ee}$,
(ii) tactile marker displacement field $\mathbf{u}$, and
(iii) previous action $a_{t-1}$.
No object pose or camera information is provided.

\noindent
\textbf{Action.}
The action represents an incremental Cartesian update of the end-effector pose, 
which is executed by the impedance controller.
This formulation ensures smooth and physically consistent motion.

\noindent
\textbf{Reward.}
The reward consists of three components:
(1) a keypoint alignment term encouraging peg–socket alignment,
(2) sparse bonuses for engagement and successful insertion, and
(3) a contact-force penalty to discourage unstable interactions.
\begin{equation}
    r_{\text{total}} = r_{\text{keypoints}} + r_{\text{engage/success}} - r_{\text{contact}}
\end{equation}

\cref{tab:random} illustrates the randomization range of parameters used in online RL training. The initial environment settings, such as controller gains, holding object initial pose and Franka hand initial pose, are randomized at episode level. The observations, containing end-effector pose and markers displacements fields, are randomized at each timestep. The randomization of markers displacements fields is performed by setting a small random movement on holding object.  \cref{tab:hyperparams} shows the detailed hyperparameters used in RL training.

\setlength{\tabcolsep}{8pt}      % Increases horizontal space between columns (default is 6pt)
\begin{table}[h]
\centering
\footnotesize   % 选项: \small (稍大), \footnotesize (推荐), \scriptsize (最小)
\caption{RL Main Hyperparameters}
\label{tab:hyperparams}
\begin{tabular}{cc}
\toprule
\textbf{Parameter} & \textbf{Value}  \\
\midrule
Learning Rate Schedule &  Adaptive  \\ % [4pt] adds 
Horizon Length &  256  \\
Minibatch Size &  512  \\
Mini Epochs &  4  \\
Sequence Length &  128  \\
RNN Type &  LSTM  \\
RNN Layers &  2  \\
RNN units &  1024  \\
MLP units &  512, 128, 64  \\
MLP activation &  ELU  \\
\bottomrule
\end{tabular}
\end{table}

\section{Real-world Deployment Results}
\begin{figure}[htbp]
  \centering
  \includegraphics[width=0.9\textwidth]{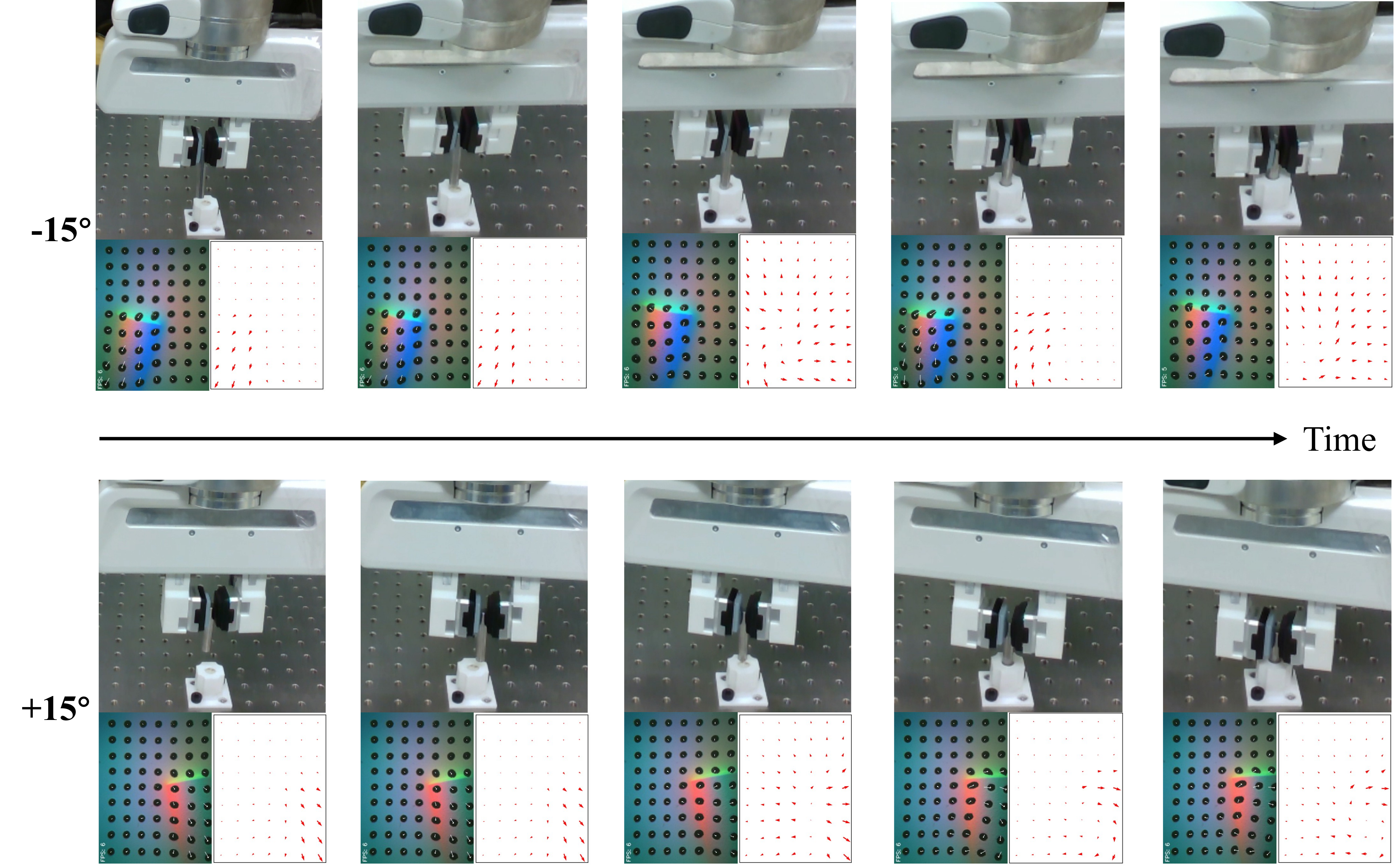}  % 替换为你的 PDF 文件名
  \caption{Supplementary real-world deployment results.}
  \label{fig:real}
\end{figure}
As for the real-world deployment setting, the socket is fixed on breadboard, and the peg is vertically put on the breadboard. We first let Franka gripper grasp the vertical peg with specific end-effector orientation, then we send the end-effector to initial inference pose and the peg with specific orientation is obtained. FrankaPy \cite{zhang2020modular} is used for real-world Franka control, and all actions are applied to end-effector. Then pose of end-effector can be read directly from FrankaPy. We also select 5 frames of the zero-shot real-world deployment for both $+15^{\circ}$ and $-15^{\circ}$, as shown in \cref{fig:real}. We found the end-effector can adjust its pose according to different peg orientation appropriately, which benefits from the tactile feedback under such "pure blind" observation setting. Furthermore, the end-effector also learns to find the correct insertion direction through the markers displacement fields under collision. These results further demonstrate the importance of tactile markers displacements field under such contact-rich task, especially for the holding object pose and collision mode identification, which offers valuable information to facilitate the inference of robot.

% \section*{Acknowledgements}
% Please insert your acknowledgments here.


\end{document}